\documentclass[10pt,twocolumn,letterpaper]{article}

\usepackage{iccv}              % To produce the CAMERA-READY version
\usepackage{mycommands}
\usepackage{mydefs}
\usepackage{wrapfig}
\usepackage{graphicx}
\usepackage{booktabs}
\usepackage{color, colortbl}
\usepackage{amsmath}
\usepackage{comment}
\usepackage{amssymb}
\usepackage{algorithm, algorithmic}
\usepackage{tikz}
\usetikzlibrary{spy}
\usepackage[accsupp]{axessibility}  % Improves PDF readability for those with disabilities.
\def\x{{\mathbf x}}

\def\sota{\emph{state-of-the-art}\xspace}
\newcommand{\et}{\emph{et al.}}
\definecolor{cvprblue}{rgb}{0.21,0.49,0.74}
\definecolor{lblue}{rgb}{0.9,0.95,1}
\definecolor{lpurple}{rgb}{0.35,0.25,0.55}
\definecolor{lgreen}{rgb}{0.95,1,0.95}
\definecolor{sblue}{rgb}{0,0.45,1}
\definecolor{lgray}{rgb}{0.95,0.95,0.95}
\definecolor{cvprblue}{rgb}{0.21,0.49,0.74}
\usepackage[pagebackref,breaklinks,colorlinks,allcolors=cvprblue]{hyperref}

\def\paperID{} % *** Enter the Paper ID here
\def\confName{ICCV}
\def\confYear{2025}

\title{Extreme Compression of Adaptive Neural Images}

\author{
}

\author{
Leo Hoshikawa$^{1*}$, Marcos V. Conde$^{1*}$, Takeshi Ohashi$^2$, Atsushi Irie$^2$ \\ \\
$^1$~Sony Interactive Entertainment \quad
$^2$~Sony Group Corporation\\
}

\begin{document}

\maketitle

%%%%%%%%%%%%%%%%%%%%%%%%%%%%%%%%%%%%%%%%%%%%%%%%%%%%%%%%%%%%%%%%%%%%%%%%

\begin{abstract}
Implicit Neural Representations (INRs) and Neural Fields are a novel paradigm for signal representation, from images and audio to 3D scenes and videos. The fundamental idea is to represent a signal as a continuous and differentiable neural network. This new approach poses new theoretical questions and challenges. Considering a neural image as a 2D image represented as a neural network, we aim to explore novel neural image compression. In this work, we present a novel analysis on compressing neural fields, with focus on images and introduce Adaptive Neural Images (ANI), an efficient neural representation that enables adaptation to different inference or transmission requirements. Our proposed method allows us to reduce the bits-per-pixel (bpp) of the neural image by 8 times, without losing sensitive details or harming fidelity. Our work offers a new framework for developing compressed neural fields. We achieve a new state-of-the-art in terms of PSNR/bpp trade-off thanks to our successful implementation of 4-bit neural representations. 
\end{abstract}

\begin{figure*}[!ht]
    \centering
    \begin{tabular}{cccc}
         \includegraphics[width=0.22\linewidth]{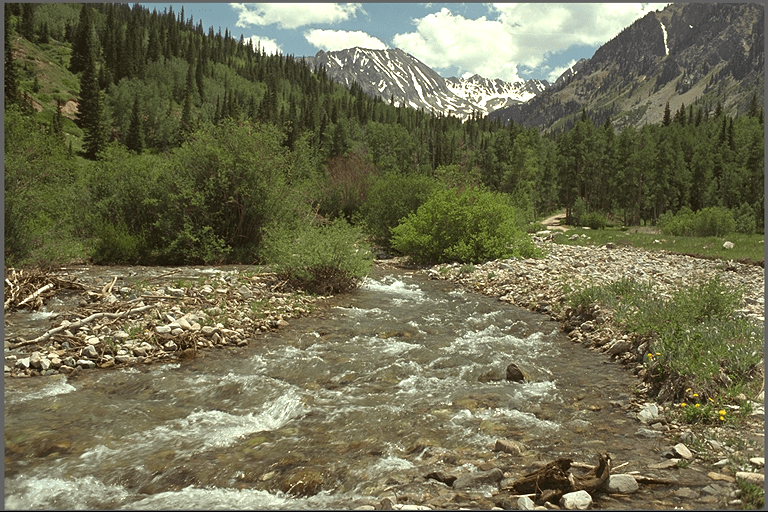} &
         \includegraphics[width=0.22\linewidth]{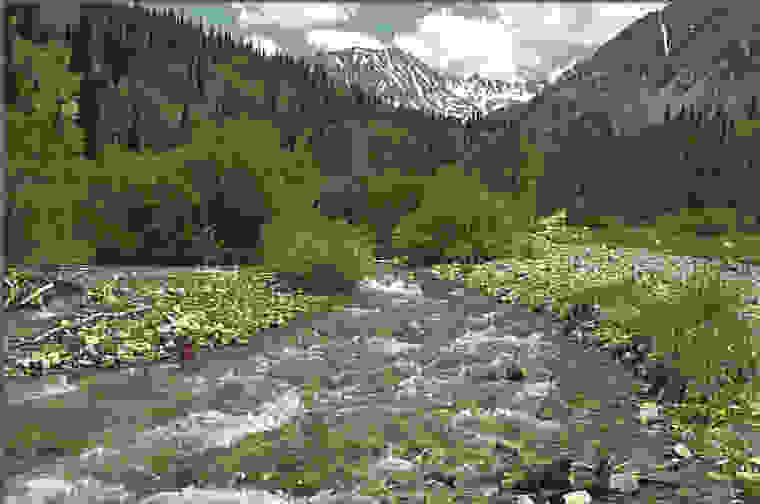} &
         \includegraphics[width=0.22\linewidth]{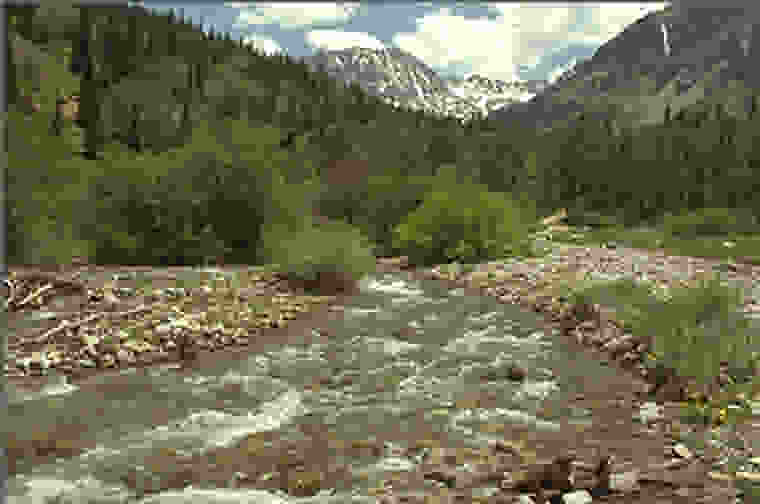} &
         \includegraphics[width=0.22\linewidth]{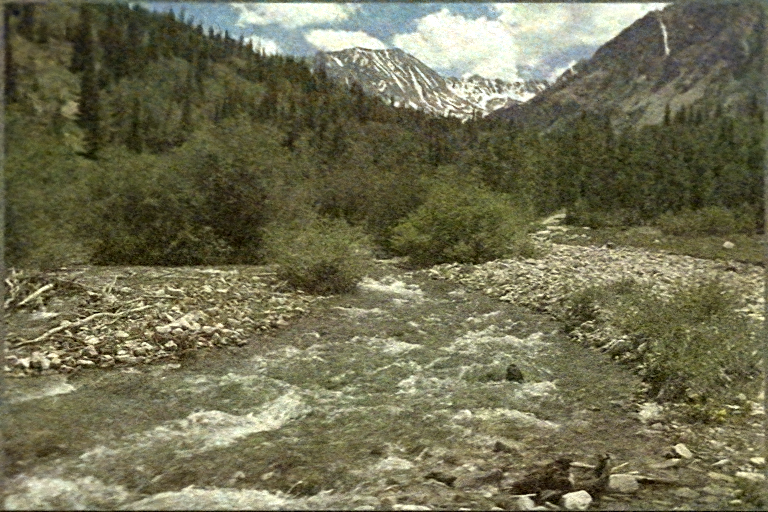} \\
         \includegraphics[width=0.22\linewidth]{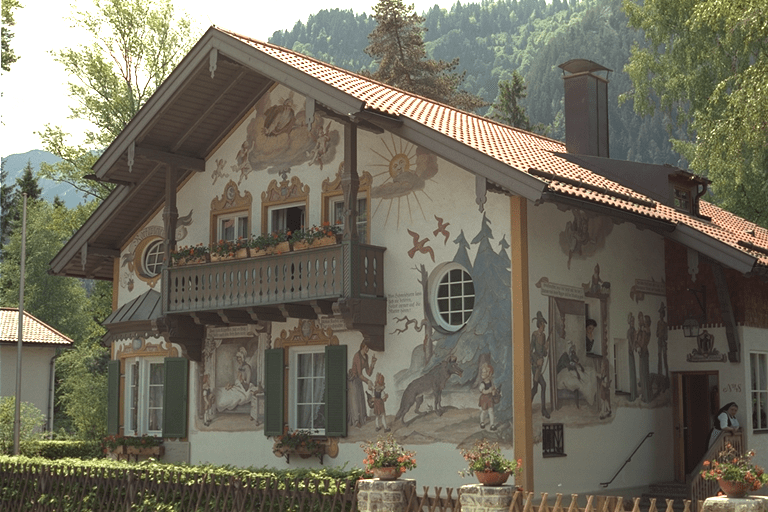} &
         \includegraphics[width=0.22\linewidth]{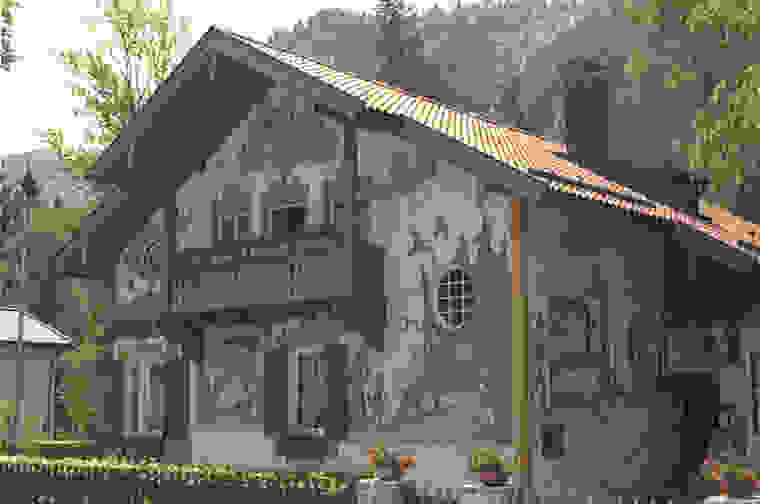} &
         \includegraphics[width=0.22\linewidth]{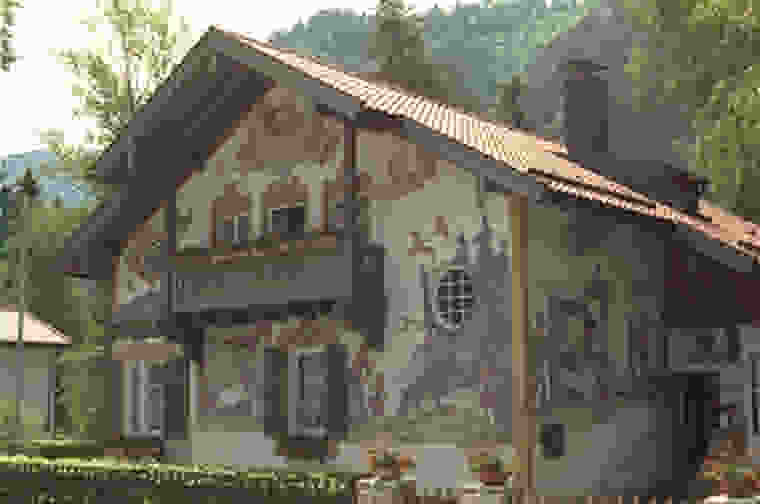} &
         \includegraphics[width=0.24\linewidth]{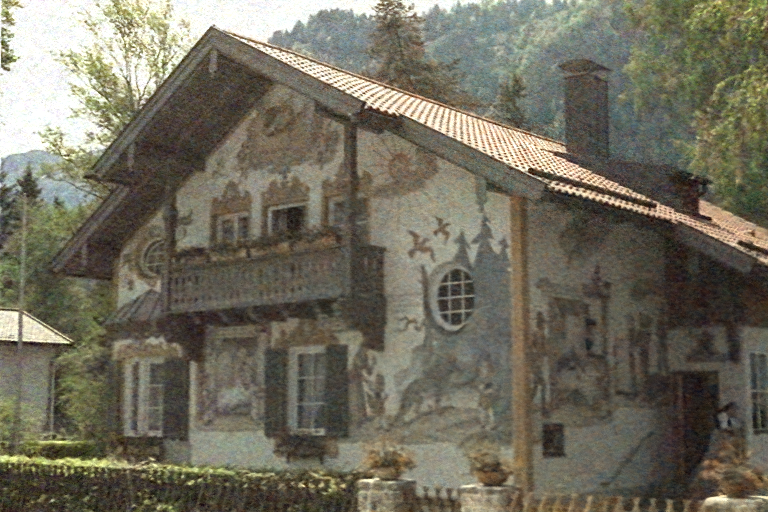} \\
         Original & JPEG & Learned JPEG~\cite{strumpler2020learning} & ANI-MFN (4-bit) \\
    \end{tabular}
    \caption{\textbf{Comparison with traditional codecs.} Our proposed neural image ANI (at 4-bits) state-of-the-art, high-fidelity results without clearly unpleasant artifacts. Note that all the images are $\approx0.3$ bpp. Images taken from the Kodak dataset IDs 13 and 24.}
    \label{fig:final_comp}
\end{figure*}

\section{Introduction}
\label{sec:intro}

Neural Fields, also known as Implicit Neural Representations (INRs), allow the representation of signals (or data) of all kinds and have emerged as a new paradigm in the field of signal processing, neural compression, and neural rendering~\cite{tancik2020fourier, sitzmann2020implicit, mildenhall2021nerf, dupont2021coin, strumpler2022implicit}. 
Unlike traditional discrete representations (\eg, image as a discrete grid of pixels, audio signals are discrete samples of amplitudes), neural fields are continuous functions that describe the signal. Such a function maps the source domain $\mathcal{X}$ of the signal to its characteristic values $\mathcal{Y}$. It maps 2D pixel coordinates to their corresponding RGB values in the image $\mathcal{I}[x,y]$. This function $\phi$ is approximated using neural networks (NNs), thus it is continuous and differentiable. 
We can formulate Neural Fields as:

\begin{equation}
    \phi : \mathbb{R}^{2} \mapsto \mathbb{R}^{3} \quad \mathbf{x} \to \phi(\mathbf{x}) = \mathbf{y},
\label{eq:inr}
\end{equation}
where $\phi$ is the learned INR function, the domains $\mathcal{X} \in \mathbb{R}^{2} $ and $\mathcal{Y} \in \mathbb{R}^{3}$, the input coordinates $\mathbf{x}=(x,y)$, and the output RGB value $\mathbf{y} = [r,g,b]$. In summary, neural representations are essentially simple neural networks (NNs), once these networks $\phi$ (over)fit the signal, the model become implicitly the signal itself.

This approach has become foundational research in many areas including image compression~\cite{dupont2021coin, strumpler2022implicit, dupont2022coin++}, audio compression~\cite{szatkowski2022hypersound, su2022inrasaudio}, video compression~\cite{chen2021nerv, chen2022videoinr} and 3D representations (\eg, NeRF, DeepSDF)~\cite{mildenhall2021nerf, muller2022instant, peng2021neural, Park_2019_CVPR}.

In the context of \emph{image compression}, this method offers unique mathematical properties due to its continuous and differentiable nature~\cite{dupont2021coin, strumpler2022implicit, dupont2022coin++}. 
One of the major advantages of using INRs is that there are no ties with spatial resolution; unlike conventional methods where the image resolution is tied to the discrete number of pixels, the memory needed for these representations only scales with the complexity of the underlying signal~\cite{tancik2020fourier, sitzmann2020implicit}. In essence, INRs offer ``infinite resolution'', it can be sampled at any spatial resolution~\cite{sitzmann2020implicit} by upsampling the input domain $\mathcal{X}$ (\eg $[H,W]$ grid of coordinates), being particularly useful for high-dimensional signal parametrization whereas traditional methods struggle due to memory limitations.

\begin{figure}[t]
     \centering
     \begin{subfigure}[b]{0.45\linewidth}
         \centering
         \includegraphics[width=\linewidth]{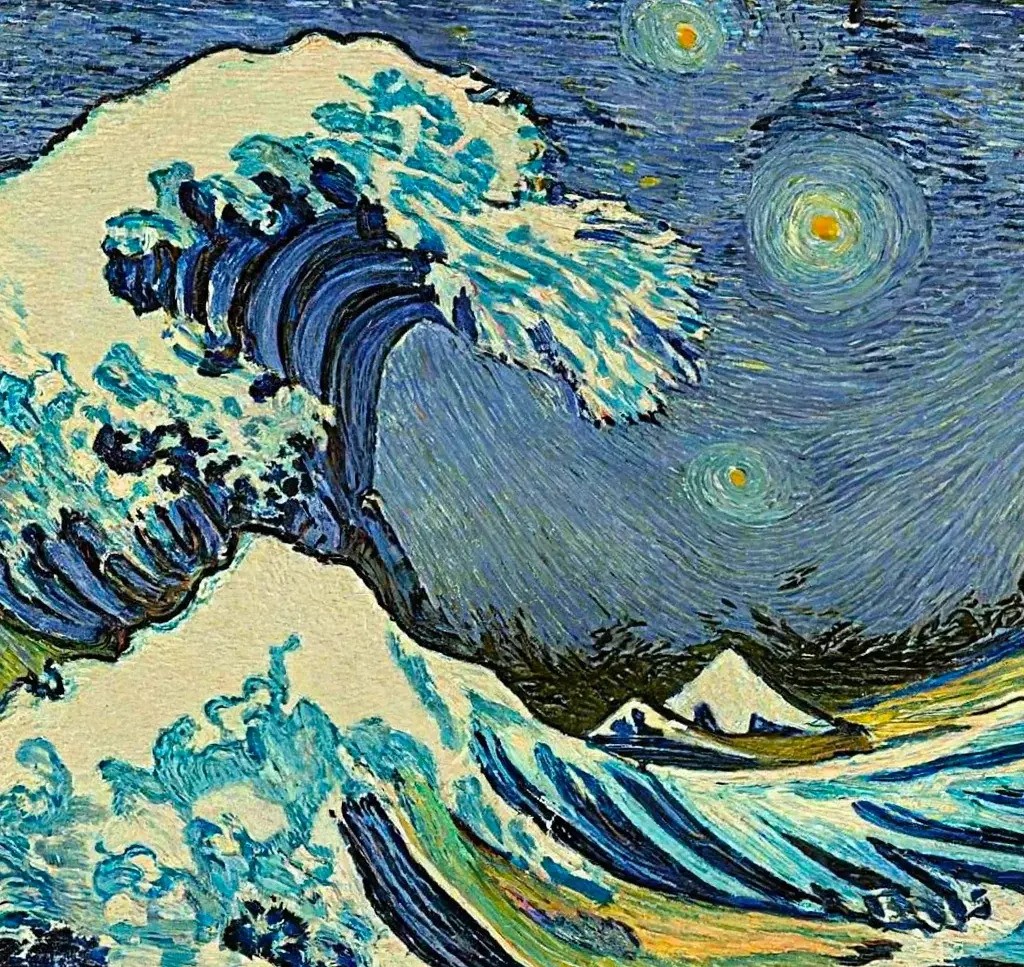}
         \put(-90,110){$\mathbf{x}= (x,y) \in [H,W]$}
         %\put(-85,130){$(x,y) \to (r,g,b)$}
         \put(-90,125){$\mathbf{x} \to (100,25,34)$}
         \put(-75,33){\textcolor{red}{$\mathbf{x}$}}
         \caption{Conventional RGB image}
         \label{fig:stream-rgb}
     \end{subfigure}
     \hspace*{5mm}
     \begin{subfigure}[b]{0.4\linewidth}
         \centering
         \includegraphics[width=\linewidth]{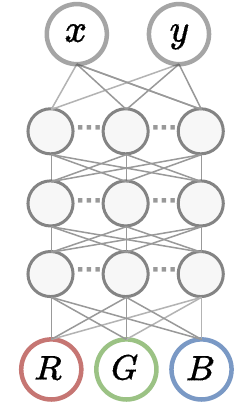}
         \caption{Coordinate-based MLP}
         \label{fig:mlp}
     \end{subfigure}
     
\vspace{-1mm}
\caption{We illustrate the general concepts around neural image representations~\cite{tancik2020fourier, sitzmann2020implicit}. INRs can be generalized to other sorts of signals such as audio or 3D representations.
}
\label{fig:inr}
\end{figure}

Considering this, we define a \emph{neural image} as a neural network (INR) that represents a particular image of an arbitrary resolution --- see Figure~\ref{fig:inr}.

Recent works~\cite{dupont2021coin, dupont2022coin++, muller2022instant} demonstrate that we can fit large images (even giga-pixel images) using ``small'' neural networks as INRs, which implies promising compression capabilities~\cite{dupont2021coin, dupont2022coin++}. %These seminal works~\cite{dupont2021coin, strumpler2022implicit} show that INRs can be a better option than standard image codecs such as JPEG~\cite{pennebaker1992jpeg} and JPEG2000~\cite{skodras2001jpeg} in the following scenarios: (i) high-resolution images, (ii) extreme compression at low bits-per-pixel (bpp). 
However, neural fields represent a \emph{lossy compression} technique, especially limited by Shannon's Theorem~\cite{shannon}; \ie even utilizing complex deep neural networks, to parameterize the high-frequencies of certain images remains a challenging or impossible task.

In this work, we focus on the particular case of 2D images, since it is well-known that this serves as a good proxy for 3D research~\cite{sitzmann2019scenes, sitzmann2020implicit, tancik2020fourier, mildenhall2021nerf}.

\paragraph{Contributions} (i) We provide an extensive benchmark on extreme image compression using neural fields. (ii) We propose Adaptive Neural Images (ANI), a novel neural representation that allows adaptation to different memory and inference requirements. We achieve this by using \sota neural architecture search (NAS) to find the optimal neural network. Our approach allows us to reduce $8\times$ the required bits-per-pixel without losing much fidelity while establishing a new state-of-the-art in PSNR/bpp ratio. (iii) We provide useful insights related to the quantization of neural fields, that can be applied to other related tasks (\ie 3D NeRF).

%%%%%%%%%%%%%%%%%%%%%%%%%%%%%%%%%%%%%%%%%%%%%%%%%%%%%%%%%%%%%%
%%% TEASER INR
%%%%%%%%%%%%%%%%%%%%%%%%%%%%%%%%%%%%%%%%%%%%%%%%%%%%%%%%%%%%%%

\section{Related Work}
\label{sec:relwork}

\noindent\textbf{Learned Image Compression.} The concept of learned image compression was pioneered by~\cite{ball2016endtoend}, through the introduction of an end-to-end framework combining an auto-encoder with an entropy model to jointly optimize both rate and distortion metrics. Many approaches~\cite{balle2018variational, minnen2018-neurips,mentzer2018conditional,lee2019context} enhanced this model by incorporating a scaling hyper-prior to the architecture, and the use of autoregressive entropy models. The current trend on generative image compression represents the state-of-the-art in terms of perceptual quality~\cite{agustsson2019generative,mentzer2020high, hoogeboom2023high, agustsson2023multi}.

\vspace{2mm}

\noindent\textbf{Model Compression.} Due to the industry requirements in terms of inference speed, memory, and energy consumption, in recent years there has been plenty of research on model compression~\cite{menghani2021efficient, xiao2023smoothquant, dettmers2024qlora}. For instance,\cite{DBLP:journals/corr/HanMD15} proposes a simple framework: applying pruning, quantization, and entropy coding --in sequence-- combined with retraining in between the steps. 
To optimize performance under quantization, several works use mixed-precision quantization and post-quantization optimization techniques~\cite{haq, Dong2019HAWQHA,uhlich2019mixed, esser2019learned, bhalgat2020lsq+, choukroun2019lowq,nagel2020down, chai2021quantizationguided}.

In particular, we adopted LSQ~\cite{esser2019learned} as the base of our quantization method. It is a strong method that improves quantization using learnable scaling factors enabling extreme low-precision settings. %In this work we will explore quantization-aware training (QAT) and post-training Quantization (PTQ).
In the context of neural fields, the neural network represents the data itself, thus, a model compression implies (additional) data compression. Despite this being a promising approach, very few works tackle this problem~\cite{gordon2023quantizing, sitzmann2020implicit, dupont2021coin, dupont2022coin++}.

\vspace{2mm}

\noindent\textbf{Neural Architecture Search (NAS)} In recent years, NAS has emerged as a powerful approach for automating the design of optimal neural network architectures for a given task, significantly reducing the need for manual experimentation~\cite{zoph2016neural, zoph2018learning}.  
The field has since seen rapid progress, with methods like Efficient Neural Architecture Search (ENAS) by Pham et al. \cite{pham2018efficient}, which significantly reduces search time by sharing weights among different architectures. 
Once-for-All~\cite{cai2019once} allows us to train a single neural network and specialize it for efficient deployment.

\subsection{Neural Representations}

In recent years, implicit neural representations (INRs)~\cite{sitzmann2020implicit, genova2019learning, muller2022instant, dou2023multiplicative} have become increasingly popular in image processing as a novel way to parameterize an image. Also known as coordinate-based networks, these approaches use multilayer perceptrons (MLPs) to overfit one image and represent it. Multiple works have demonstrated the potential of MLPs as continuous, memory-efficient implicit representations for images~\cite{sitzmann2020implicit, strumpler2022implicit}.

We denote the INRs as a function $\phi$ with parameters $\theta$, defined as: 

\begin{equation} \label{eq:mlp}
\begin{split}
\phi (\mathbf{x}) = \mathbf{W}_n ( \varsigma_{n-1} \circ \varsigma_{n-2} \circ \ldots \circ \varsigma_0 )(\mathbf{x}) + \mathbf{b}_n \\
\varsigma_i (\mathbf{x_i}) = \alpha \left( \mathbf{W}_i \mathbf{x}_i + \mathbf{b}_i \right),
\end{split}
\end{equation}

where $\varsigma_i$ are the layers of the network (considering their corresponding weight matrices $\mathbf{W}$ and bias $\mathbf{b}$), and $\alpha$ is a nonlinear activation \eg ReLU, Tanh, Sine~\cite{sitzmann2020implicit}, complex Gabor wavelet~\cite{saragadam2023wire}. Considering this formulation, the parameters of the neural network $\theta$ is a set of weights and biases of each layer (\ie $\mathbf{W}$ and $\mathbf{b}$). Since the input of the MLP are the coordinates $\mathbf{x}$ in the domain $[H,W] \in \mathbb{R}^{2} $, these are also known as coordinate-based MLPs --- see Figure~\ref{fig:mlp}.

Sitzmann \et~\cite{sitzmann2020implicit} presented SIREN, a periodic activation function for neural networks based on the Sine function, specifically designed to better model complex natural signals and high-frequencies in the images. Tancik \et~\cite{tancik2020fourier} introduced Fourier features as input positional encodings for the network, enhancing their capability to model high-frequencies. COIN~\cite{dupont2021coin, dupont2022coin++} explored the early use of INRs for image compression. Strumpler~\et~\cite{strumpler2022implicit} proposed a framework for image compression and transmission using INRs. We also find other works that tackle new activation functions such as multiplicative filter networks (MFN)~\cite{fathony2020multiplicative} and Wire~\cite{saragadam2023wire}, and multi-scale representations~\cite{saragadam2022miner, muller2022instant}. Other such as Instant-NGP~\cite{muller2022instant} and SHACIRA~\cite{girish2023shacira} approaches focus on multi-resolution representations using hierarchical representations and hash-tables to improve performance and speed.

Following previous work~\cite{dupont2021coin, strumpler2022implicit}, we will use SIREN~\cite{sitzmann2020implicit} as the baseline model. We will explore extreme compression of the neural network, and new training techniques to derive our proposed adaptive neural images (ANI). We will also analyze the most popular and recent approaches: FourierNets~\cite{tancik2020fourier} (MLP with Positional Encoding), SIREN~\cite{sitzmann2020implicit}, MFN~\cite{fathony2020multiplicative}, Wire~\cite{saragadam2023wire} and DINER~\cite{xie2023diner}.

%%%%%%%%%%%%%%%%%%%%%%%%%%%%%%%%%%%%%%%%%%%%%%%%%%%%%%%%%%%%%%
%%%% STREAMING FIGURE

\begin{figure*}[ht!]
    \centering
    
    \begin{subfigure}[b]{0.9\linewidth}
        \includegraphics[width=\linewidth]{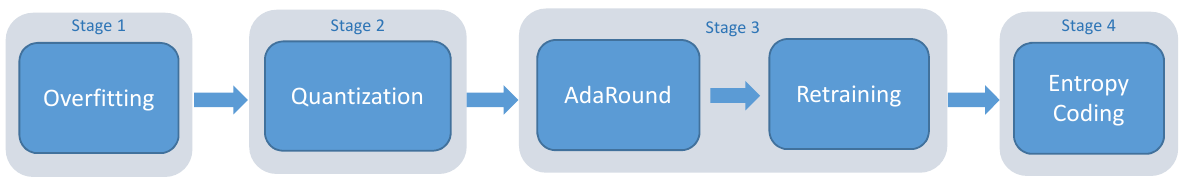}
        \caption{Overview of INR-based compression pipeline proposed by Y. Strümpler et al.~\cite{strumpler2022implicit}. The basic compression pipeline comprises image overfitting, quantization of the neural network, AdaRound, retraining, and lossless entropy coding (\eg binarized arithmetic coding).}
        \label{fig:inr_compresion}
    \end{subfigure}
    
    \vspace{0.5cm}
    
    \begin{subfigure}[b]{\linewidth}
        \includegraphics[width=0.9\linewidth]{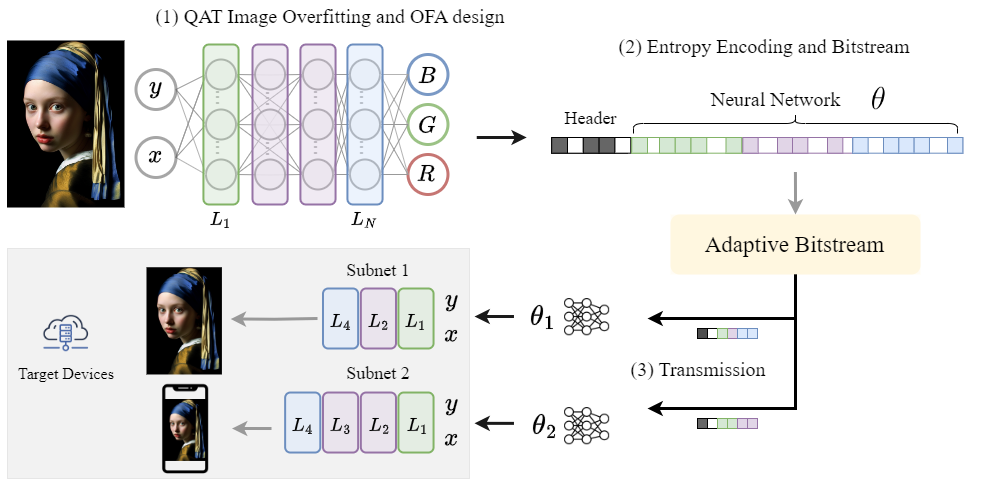}
        \vspace{1mm}
        \caption{Our proposed approach uses adaptive neural images (ANI). We perform directly quantization aware training (QAT)~\cite{bhalgat2020lsq+} and once-for-all (OFA) optimization~\cite{cai2019once}. Depending on the bandwidth and transmission requirements, our bitstream can be adapted (\eg trimmed) allowing us to send more/less information, this is only possible thanks to the proposed ANI architecture. Moreover, depending on the target device speed and memory requirements, we can utilize smaller versions of our neural network ANI without any re-training or adaptation. We highlight the \colorbox{lgray}{client side} and target devices.
        }
        \label{fig:our_pipe}
    \end{subfigure}
    
    \caption{We illustrate the general concepts around image compression and transmission using INRs~\cite{strumpler2022implicit}. Our approach enables to adapt to diverse scenarios depending on the bandwidth, memory, and target device requirements.}
    \label{fig:streaming}
\end{figure*}

%%%%%%%%%%%%%%%%%%%%%%%%%%%%%%%%%%%%%%%%%%%%%%%%%%%%%%%%%%%%%%

\section{Transmission of Neural Images}

Transmitting signals as INRs is a novel research problem~\cite{strumpler2022implicit, dupont2022coin++}. In this context, it is fundamental to understand that the image is no longer characterized as a discrete set of RGB pixels, but as a set of weights and biases ($\theta$ \ie the neural network itself). 
We illustrate in Figure~\ref{fig:inr_compresion} the most popular approach for compressing images using INRs. First, we train the neural network $\phi$ to fit the image, next we can apply post-training quantization (PTQ) and encode the parameters $\theta$ using lossless entropy coding. We could also apply post-quantization retraining to improve the performance of the neural network. Finally, we can transmit the parameters $\theta$, the client can recover the network, and thus reconstruct the natural RGB image. Our approach considers quantization-aware training (QAT) directly, which offers better performance and a higher compression ratio. We show our method in Figure~\ref{fig:our_pipe}.

Besides QAT, the key concept of our approach is the active neural architecture search (NAS) to produce a ``once-for-all'' neural network~\cite{cai2019once} \ie a single network is trained to support versatile architectural configurations including depth (number of layers) and width (number of neurons). Therefore considering our neural image with parameters $\theta$ we can derive --during inference-- different sub-networks with varying number of layers and neurons. We illustrate the sub-networks $\theta_1$ and $\theta_2$ in Figure~\ref{fig:our_pipe}.

\vspace{6mm}
We define \emph{\bf Adaptive Neural Images (ANIs)} as once-for-all neural representations of images. Note that ANIs are also trained to support quantization. Note that the neural representation training is done \emph{offline} only once for a particular image, thus, the training time is not constraint. Moreover, training to convergence is possible in a few minutes.

\vspace{2mm}
\noindent\textbf{General Limitations} Before presenting our approach, we must discuss the fundamental limitations of neural images to better understand the experimental results. First, INRs are lossy compression methods. Second, most INR approaches are signal-specific \ie the neural network fits a particular image. This implies training \emph{ad hoc} the neural network using a GPU --- although this can take less than 1 minute, and meta-learning~\cite{tancik2021metalearned} helps to accelerate training. Third, the performance of the INR methods highly varies depending on hyper-parameters (\eg learning rate, number of neurons and layers), and the target signal. However, there is no theoretical
or practical way of predicting \emph{a priori} which INR model fits best the signal.

%%%%%%%%%%%%%%%%%%%%%%%%%%%%%%%%%%%%%%%%%%%%%%%%%%%%%%%%%%%%%%%%

\section{Our Approach for Extreme Compression}
\label{sec:ours}

Given a neural representation of an image --a neural image--, our goal is to reduce as much as possible the number of bits while preserving the original signal. Considering that the neural network represents the signal itself, we must focus on compressing the neural network (\ie weights and biases).

%%%%%%%%%%%%%%%%%%%%%%%%%%%%%%%%%%%%%%%%%%%%%%%%%%%%%%%%%%%%%%%

\subsection{Post-Training Quantization}
\label{sec:post-q}
Post-Training Quantization (PTQ) calculates quantization parameters without re-training. In our experiments, we adopted the standard PTQ algorithm proposed by~\cite{jacob2018ptq} and wide used on several studies~\cite{strumpler2022implicit}. This algorithm allows quantization to 7-bits and 8-bits with minimal losses.

\subsection{Quantization-Aware Training (QAT)}
\label{sec:qat}

Quantization-aware training (QAT) methods have a considerable advantage over PTQ methods in terms of compression ratio~\cite{bhalgat2020lsqplus, DBLP:journals/corr/HanMD15, esser2019learned, jain2020trained}, allowing extremely small bit-width (2, 4-bit) at the expense of additional training time. 

In general neural networks, weights follow zero-mean normal distributions, while the distribution for the activations varies greatly depending on the architecture and non-linearities. 
For INRs, the behavior of the MLP and activations is well-known. In SIREN~\cite{sitzmann2020implicit} the sine activation conveniently restricts the distribution to a normalized range with zero-mean. For MFNs~\cite{fathony2020multiplicative}, the filter passes through the sine activation and is multiplied by the output of the linear layers, conveniently restricting the range. These properties allow us to experiment with extreme compression low-bits settings (2,4-bits).

Unlike previous methods~\cite{strumpler2022implicit, gordon2023quantizing}, we use the \sota LSQ~\cite{esser2019learned} quantization algorithm. 
\begin{comment}
We adapt this algorithm for INRs as follows:

\begin{algorithm}
\label{al:lsqp}
\caption{Quantization-Aware Training using Learnable Step Size Quantization (LSQ+)~\cite{esser2019learned}}
\begin{algorithmic}
\REQUIRE Input tensor $X$, number of bits $b$, learnable step size $s$
\ENSURE Quantized tensor $Q$

\STATE $Q_{\max} \leftarrow 2^{b-1} - 1$
\STATE $Q_{\min} \leftarrow -2^{b-1}$

\STATE // Forward pass quantization
\STATE $X_{\text{scaled}} \leftarrow X / s$
\STATE $X_{\text{clipped}} \leftarrow \max(\min(X_{\text{scaled}}, Q_{\max}), Q_{\min})$
\STATE $Q \leftarrow \text{round}(X_{\text{clipped}})$

\STATE // During backpropagation, update $s$ using gradient descent

\RETURN $Q \times s$ \COMMENT{Return de-quantized tensor for subsequent computations}

\end{algorithmic}
\end{algorithm}
\end{comment}
%%%%%%%%%%%%%%%%%%%%%%%%%%%%%%%%%%%%%%%%%%%%%%%%%%%%%%%
%%% PTQ vs QAT

\begin{figure*}[!ht]
    \centering
    \setlength{\tabcolsep}{1pt} % Default value: 6pt
    \begin{tabular}{c c c c}
        \multicolumn{2}{c}{\includegraphics[width=0.49\linewidth]{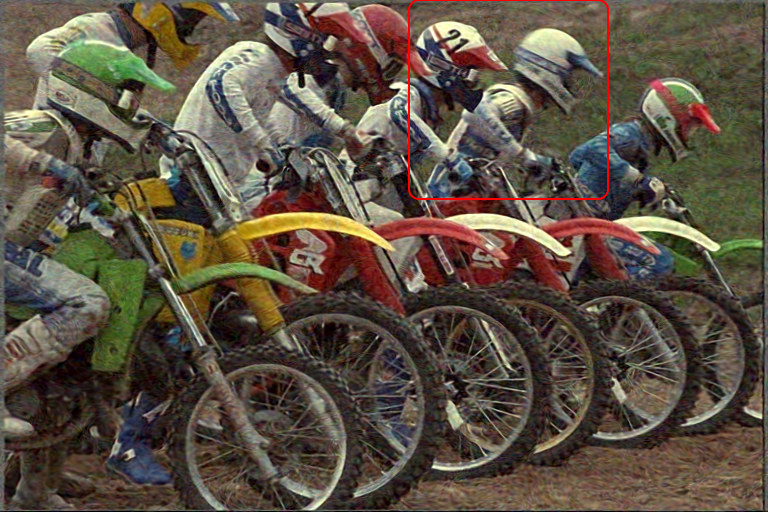}} &

        \multicolumn{2}{c}{\includegraphics[width=0.49\linewidth]
        {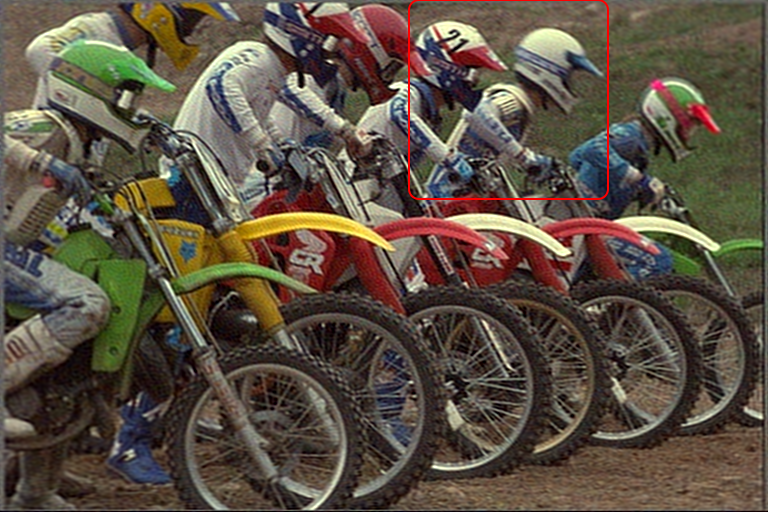}} \\

        \multicolumn{2}{c}{SIREN (Coin)~\cite{dupont2021coin}} & \multicolumn{2}{c}{MFN~\cite{fathony2020multiplicative}} \\

        % PTQ
        \includegraphics[width=0.24\linewidth]{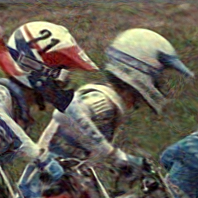} &
        \includegraphics[width=0.24\linewidth]{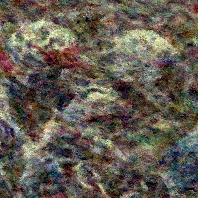} &
        \includegraphics[width=0.24\linewidth]{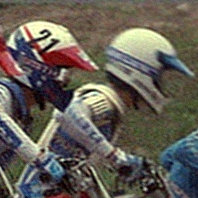} &
        \includegraphics[width=0.24\linewidth]{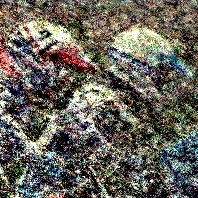} \\

        PTQ 8-bit~\cite{strumpler2022implicit} & PTQ 4-bit & PTQ 8-bit & PTQ 4-bit \\

        % QAT
        \includegraphics[width=0.24\linewidth]{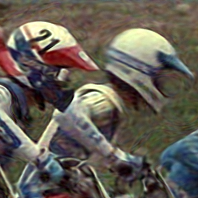} & 
        
        \includegraphics[width=0.24\linewidth]{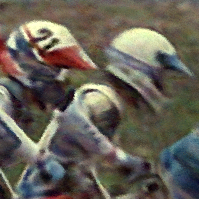}  &

        \includegraphics[width=0.24\linewidth]{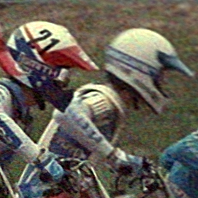} & 
        
        \includegraphics[width=0.24\linewidth]{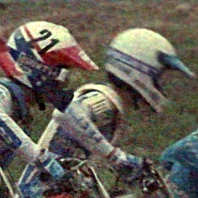}  \\

        QAT 8-bit (\emph{ours}) &  QAT 4-bit (\emph{ours}) & QAT 8-bit (\emph{ours}) &  QAT 4-bit (\emph{ours}) \\
         
    \end{tabular}
    \caption{\textbf{Comparison between PTQ and QAT.} Visual results on Kodak~\cite{kodak} at different bit-widths. We can appreciate how at 4-bits PTQ loses the signal, while QAT maintains high fidelity. Our method improves previous approaches~\cite{strumpler2022implicit, dupont2021coin, dupont2022coin++}.
    }
    \label{fig:ablation_quantize_ptq}
\end{figure*}
%%%%%%%%%%%%%%%%%%%%%%%%%%%%%%%%%%%%%%%%%%%%%%%%%%%%%%%%

Following the notation from~\cite{esser2019learned}, we define $\bar{x}$ and $\hat{x}$ as the coded bits and quantized values, respectively. The weights and activations are quantized as follows:

\begin{equation}
\bar{x}=quantize(clamp(\frac{x}{s},x_{min}, x_{max})) \quad \text{, } \quad \hat{x}=\bar{x}\times s 
\end{equation}

The \textit{s} parameters are clipping factors learned using back-propagation~\cite{esser2019learned}. For activations, $x_{min}$ is \textit{0} if the \textit{x} is strictly positive, or \textit{-1} otherwise, and $x_{max}$ is always 1. For weights, $\x_{min}$ and $x_{max}$ are always \textit{-1} an \textit{1} respectively. On \text{n}-bits, quantization is given by: 

\begin{equation}
\bar{x} = \frac{round((\bar{x}+1)\times 2^{n-1})}{2^n} \quad \text{, \textit{n}=number of bits.}
\end{equation}

The same equation is applied to both the weights and activations. We use straight-through estimator (STE)~\cite{bengio2013estimatingste} and update the quantization parameters using back-propagation. In Figure~\ref{fig:ablation_quantize_ptq} we show the benefits of using LSQ~\cite{esser2019learned} quantization-aware training (QAT) over post-training quantization (PTQ).

\subsection{Neural Architecture Search (NAS)}
\label{sec:nas}

We experimented with NAS to find optimal architectures automatically. Since we expect multiple target devices and different specifications, we use Once-for-All (OFA)~\cite{cai2020ofa}, a supernet approach that allows training once and extracting multiple sub-architectures of different sizes with minimal retraining, and adapted to INRs. Since INRs are essentially MLPs, the only moving parts that  can be made ``elastic'' are the depth (number of layers) and the width (number of channels or neurons). Additionally, to simplify the search space, we adopted a uniform number of channels for all intermediate layers. While this restricts the search space, it allows us to train and evaluate several possible layouts. 

During training, the subnets are initialized using progressive shrinking from ~\cite{cai2020ofa}. we alternate between large and small networks to remove architecture-related bias, inspired by the sandwich rule proposed by~\cite{Yu2020@bignas}. Next, we fine-tune the sub-networks for a small amount of epochs to improve the fidelity \emph{w.r.t} of the target image.

\begin{algorithm}
\caption{Once-for-all training strategy}
\begin{algorithmic}
\REQUIRE Search space of channels $W$=\{$W_0$,$W_1$,...,$W_n$\}, layers $D$=\{$D_0$,$D_1$,...,$D_n$\} 
\STATE $S$ = $W \times D$
\FOR{ each $s \in S$} 
\STATE params = get\_model\_size(s)
\ENDFOR
\STATE //Argsort S using params
\STATE $idx = argsort(params)$
\STATE //Reorder S by alternating large and small architectures
\STATE $S_{sorted} = sort\_and\_shuffle(S,idx)$
\STATE $supernet = build\_model(W_n,D_n)$ 
\STATE $train(supernet)$
\FOR{ each $s \in S_{sorted}$}
\STATE $subnet = supernet.get\_subnet(s)$
\STATE $train(subnet)$
\ENDFOR
\RETURN subnet

\end{algorithmic}
\end{algorithm}

This OFA approach allows us to realize the adaptive bitstream with minimum effort. 

\section{INR Compression Benchmark}
\label{sec:results}

\begin{table}[t]
    \resizebox{\linewidth}{!}{%
    \centering
    \begin{tabular}{l l c c c c}
        \toprule
        \rowcolor{lgray} Method & Quantization & Size(KB) & PSNR~$\uparrow$ & SSIM~$\uparrow$ & BPP~$\downarrow$ \\
        \midrule
         \multirow{5}{*}{SIREN~\cite{sitzmann2020implicit}} & Coin~\cite{dupont2021coin} (None) & 270.28 & 27.98$\pm$2.73 & 0.782 & 1.812 \\ 
         & PTQ 8-bit~\cite{strumpler2022implicit} & 71.46 & 27.78$\pm$2.71 & 0.760 & 0.479 \\
         & PTQ 4-bit (\emph{ours}) & 38.33 & 18.24$\pm$1.72 & 0.216 & 0.236 \\
         %\midrule
         & QAT 8-bit (\emph{ours}) & 71.46 & 27.80$\pm$2.34 & 0.742 & 0.479 \\
         & QAT 4-bit (\emph{ours}) & 38.33 & 27.59$\pm$3.30 & 0.638 & 0.236 \\        
         \midrule
         \midrule
         \multirow{5}{*}{MFN~\cite{fathony2020multiplicative}} & Coin~\cite{dupont2021coin} (None) & 284.29 & 29.16$\pm$2.80 & 0.822 & 1.906 \\         
         & PTQ 8-bit~\cite{strumpler2022implicit} & 85.43 & 28.60$\pm$2.82 & 0.785 & 0.572 \\
         & PTQ 4-bit (\emph{ours}) & 52.29 & 14.22$\pm$2.20 & 0.136 & 0.254 \\
         & QAT 8-bit (\emph{ours}) & 85.43 & 29.86$\pm$2.99 & 0.780 & 0.572 \\
         & QAT 4-bit (\emph{ours}) & 52.30 & 28.50$\pm$2.72 & 0.683 & 0.254 \\
         \bottomrule
    \end{tabular}}
    \vspace{2mm}
    \caption{\textbf{Quantization INR Analysis} on Kodak~\cite{kodak}. We report the average PSNR --over 5 runs-- for the whole Kodak image dataset using different quantization settings. All the neural networks have 4 layers and 128 neurons. We are the first approach to achieve successful 4-bit quantization of neural images.}
    \label{tab:quant_kodak_whole}
\end{table}

\begin{table}[t]
    \centering
    \begin{tabular}{l c c c}
        \toprule
        \rowcolor{lgray} BPP range & Bitwidth & l$\times$ch & PSNR~$\uparrow$ \\
        \midrule
         \multirow{2}{*}{0.1} & 8-bit & 3$\times$64 & 25.36$\pm$2.64 \\
         & \textbf{4-bit}  & \textbf{2$\times$128} & \textbf{26.39$\pm$2.62} \\
         \midrule
        \multirow{2}{*}{0.5} & 8-bit & 4$\times$128 & 29.86$\pm$2.98 \\
         & \textbf{4-bit}  & \textbf{2$\times$256} & \textbf{30.10$\pm$3.06} \\
         \midrule
        \multirow{2}{*}{1.0} & 8-bit & 4$\times$128 & 30.24$\pm$2.98 \\
         & \textbf{4-bit}  & \textbf{4$\times$256} & \textbf{33.10$\pm$3.06} \\
         \bottomrule
    \end{tabular}
    \caption{\textbf{Cost-efficiency of 4-bits method} on Kodak~\cite{kodak} using MFN backbone. For the same bpp budget, the 4-bit model achieves superior PSNR for all bpp ratios.}
    \label{tab:quant_kodak_same_bpp}
\end{table}
%%%%%%%%%%%%%%%%%%%%%%%%%%%%%%%%%%%%%%%%%%%%%%%%%%%
%%%%%%%%%%%%%%%%%%%%%%%%%%%%%%%%%%%%%%%%%%%%%%%%%%%
\subsection{Quantization experiments}
We run an exhaustive benchmark using the Kodak dataset~\cite{kodak}, we provide the results in Table~\ref{tab:quant_kodak_whole}. Each experiment was repeated five times with different random seeds. We report the average performance of the five experiments. In all the experiments we used models with 4 layers and 128 channels.

For the PTQ experiments, we follow~\cite{strumpler2022implicit}. Since the results for this technique are deterministic, we select the best-performing model per image (considering the five different runs). Following other quantization experiments, we kept the first and last layers at full precision. The impact of quantizing these layers is described as an ablation study. 

\subsection{Quantized NAS experiments}
To develop our adaptive neural images (ANIs) we use OFA~\cite{cai2020ofa}. Using this NAS technique, we defined our search space of $[64,128,192,256]$ channels and $[2,3,4,5]$ intermediate layers. We train all possible 16 architecture combinations for 50000 epochs each.

\vspace{2mm}

\begin{figure*}[t]
    \centering
    \begin{tabular}{c c c}        
         \rotatebox{90}{\hspace{2.1mm} ANI-MFN}
         \includegraphics[width=0.3\linewidth]{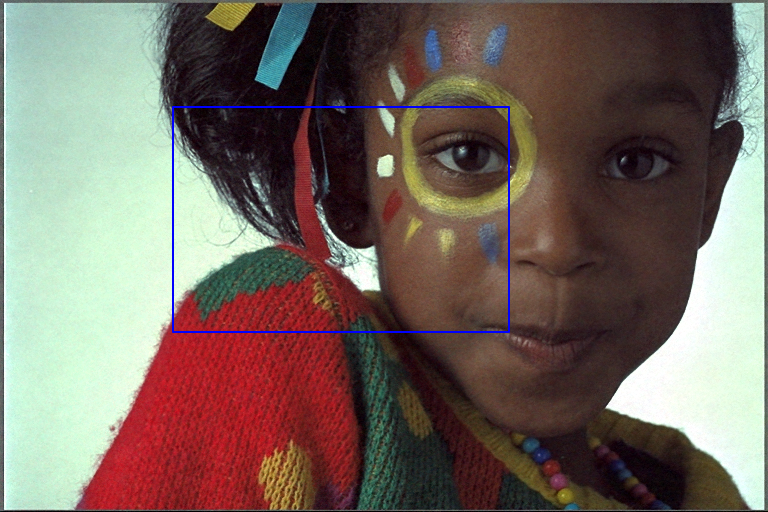} & 
         \includegraphics[width=0.3\linewidth]{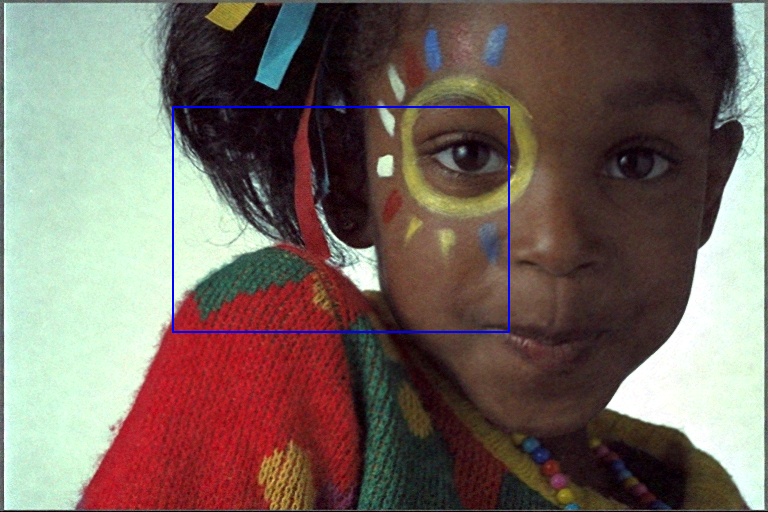} & 
         \includegraphics[width=0.3\linewidth]{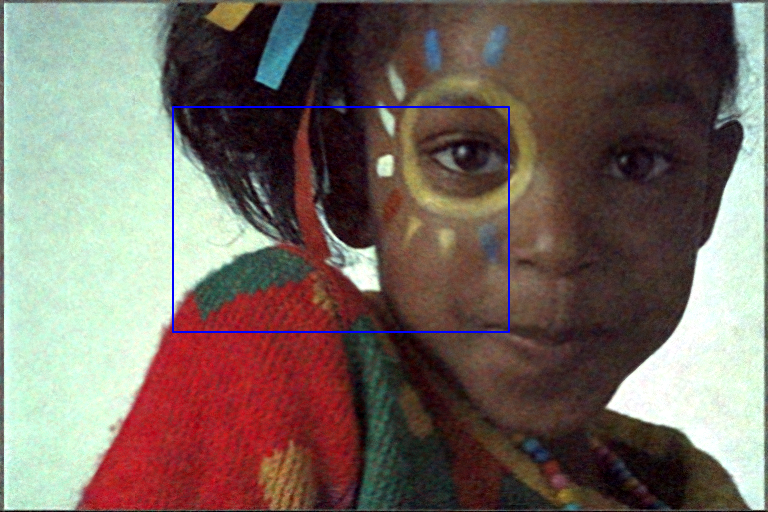} \\
         \includegraphics[width=0.3\linewidth]{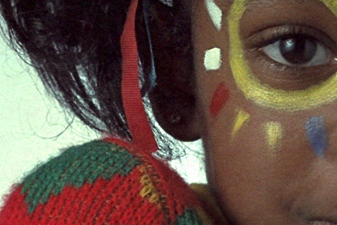} & 
         \includegraphics[width=0.3\linewidth]{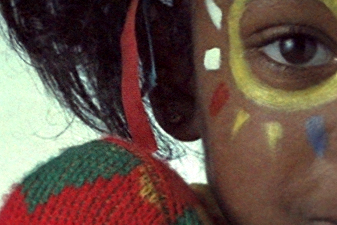} & 
         \includegraphics[width=0.3\linewidth]{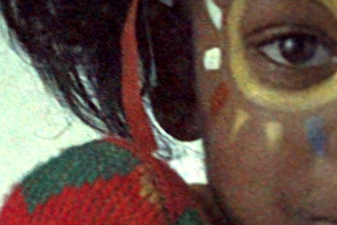} \\
         4x256 & 2x256 & 4x128 \\
         $\approx0.93$ bpp & $\approx0.47$ bpp & $\approx0.25$ bpp \\
         \midrule
         \rotatebox{90}{\hspace{2.1mm} Other methods}
         \includegraphics[width=0.3\linewidth]{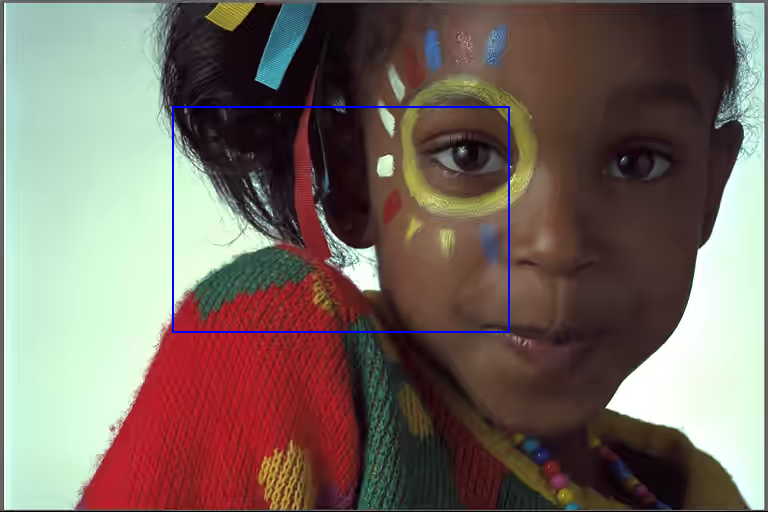} & 
         \includegraphics[width=0.3\linewidth]{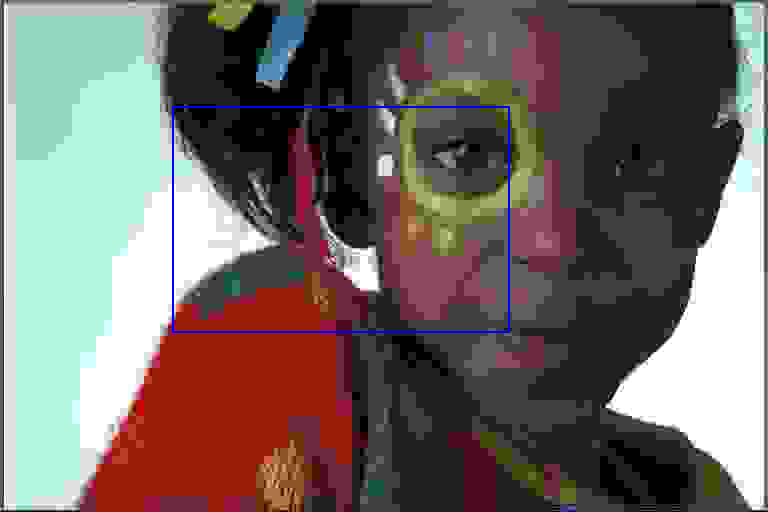} & 
         \includegraphics[width=0.3\linewidth]{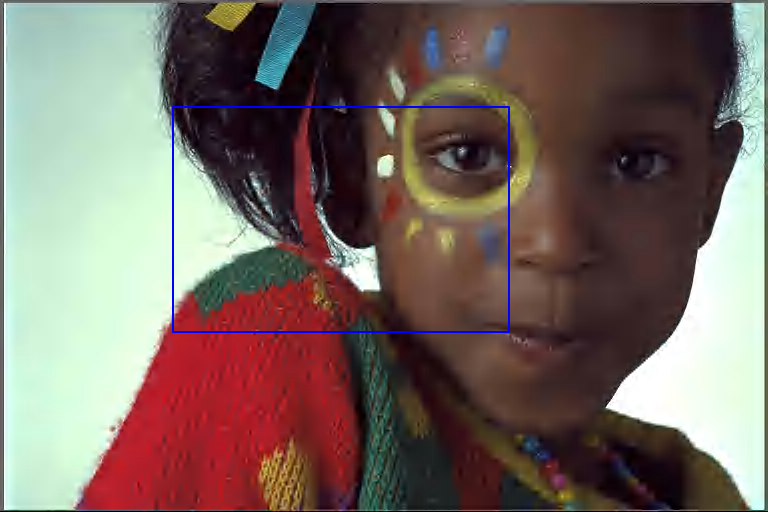} \\
        \includegraphics[width=0.3\linewidth]{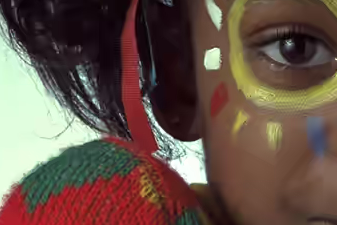} & 
         \includegraphics[width=0.3\linewidth]{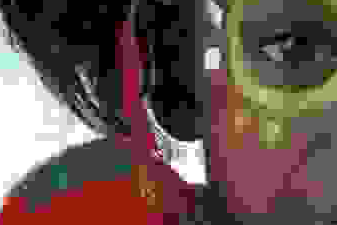} & 
         \includegraphics[width=0.3\linewidth]{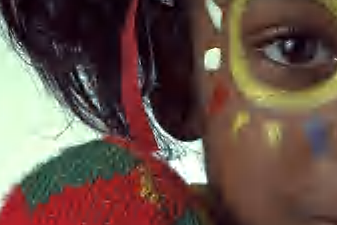} \\
         BPG & JPEG~\cite{pennebaker1992jpeg} & JPEG2000~\cite{skodras2001jpeg}
    \end{tabular}
    \caption{We present results of ANI using MFN~\cite{fathony2020multiplicative} as backbone. Our single neural image can be adapted to different memory-fidelity requirements. The images correspond to a single neural network with three different subnetworks defined as layers$\times$neurons. 
    Our methods achieve better performance than JPEG~\cite{pennebaker1992jpeg} and JPEG2000~\cite{skodras2001jpeg} at $\approx0.23$ bpp.}
    \label{fig:ofa_comp}
\end{figure*}

\noindent\textbf{Benchmark Conclusions} Considering the results from Table~\ref{tab:quant_kodak_whole}, we are the first approach to achieve successful 4-bit quantization of neural representations. At 8-bits, both PTQ and QAT deliver similar quality without notable degradations.  However, at 4-bits, the model quantized with PTQ loses the signal information, yet the model quantized with QAT maintains the signal and provides good fidelity. Our approaches improve Coin~\cite{dupont2021coin} and previous INR compression~\cite{strumpler2022implicit} by \textbf{+14dB} when using 4-bits. 
Figure~\ref{fig:ablation_quantize_ptq} shows the visual results. Both SIREN~\cite{sitzmann2020implicit} and MFN~\cite{fathony2020multiplicative} presented similar behavior during quantization, with SIREN~\cite{sitzmann2020implicit} being slightly more cost-efficient due to having fewer full precision parameters but suffers more degradation than MFN counterpart. In Figure~\ref{fig:ofa_comp}, we show the results of our ANI \ie a single super-network that allows inferring using sub-networks depending on the memory requirements. We provide more qualitative samples in the appendix.

\begin{figure}[t]
    \centering
    \includegraphics[width=\linewidth]{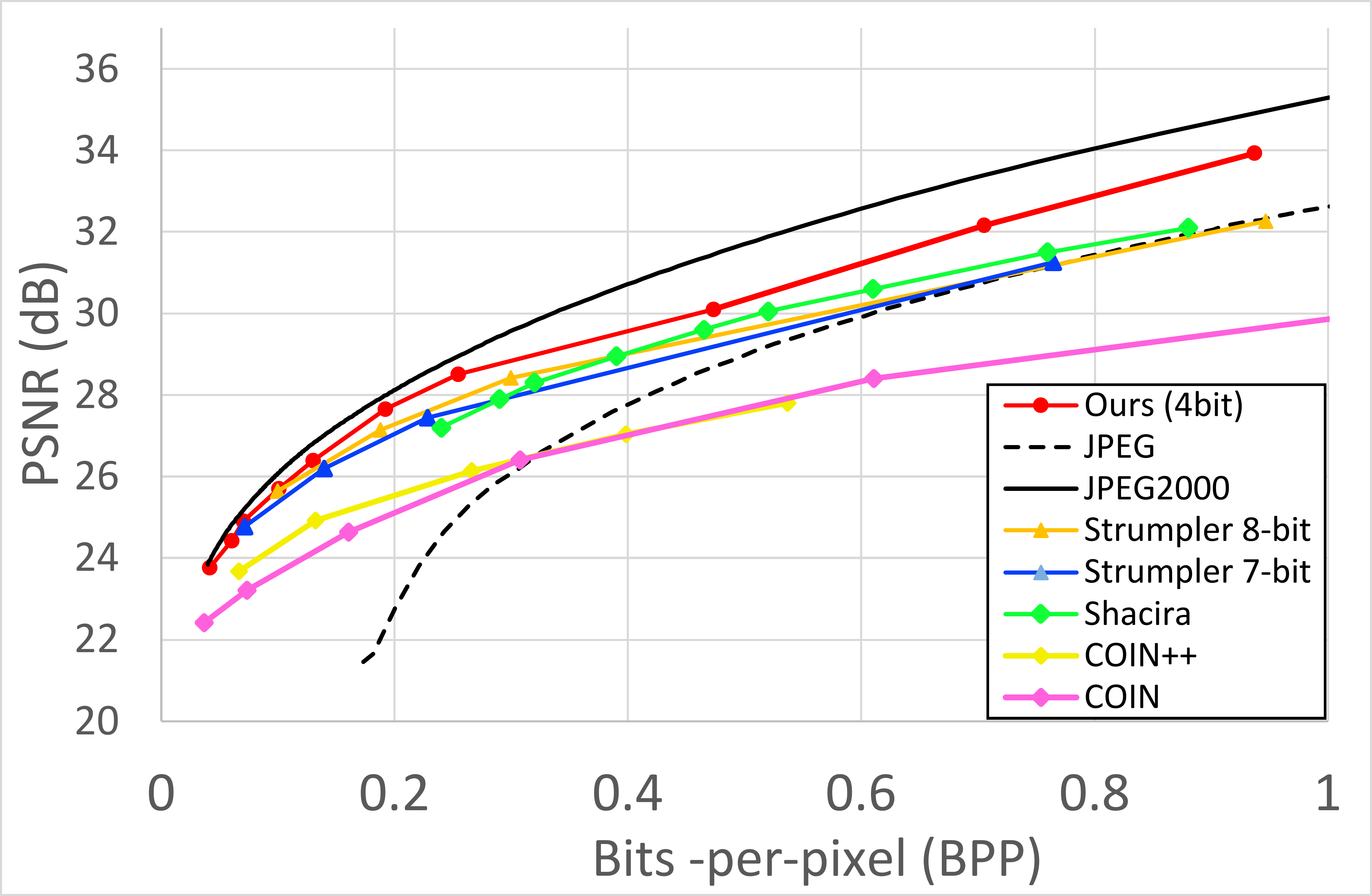}
    \caption{Comparison of our approach on the Kodak dataset with other methods. We achieved state-of-the-art performance, surpassing even newer methods such as SHACIRA~\cite{girish2023shacira}. Note that ANI-MFN is a single neural network that can be adapted to different bpp requirements, unlike Coin~\cite{dupont2021coin, dupont2022coin++} or SIREN~\cite{sitzmann2020implicit, strumpler2022implicit}.
    }
    \label{fig:kodak_baselines}
\end{figure}

We compare our approach with other compression methods in Figure~\ref{fig:kodak_baselines}. Our approach using 4-bits presents the best PSNR/ bpp trade-off along the whole spectrum, establishing a new state-of-the-art for INR compression.

%%%%%%%%%%%%%%%%%%%%%%%%%%%%%%%%%%%%%%%%%%%%%%%%%%%%%%%%%%%%
%%% PSNR-BPP PLOT
%%%%%%%%%%%%%%%%%%%%%%%%%%%%%%%%%%%%%%%%%%%%%%%%%%%%%%%%%%%%

%%%%%%%%%%%%%%%%%%%%%%%%%%%%%%%%%%%%%%%%%%%%%%%%%%%%%%%%%%%%

\subsection{NeRF Extension}

\begin{figure}[!ht]
    \centering
    \setlength{\tabcolsep}{1pt} % Default value: 6pt
    \begin{tabular}{c c c c}
         \multicolumn{4}{c}{
         \includegraphics[width=0.95\linewidth]{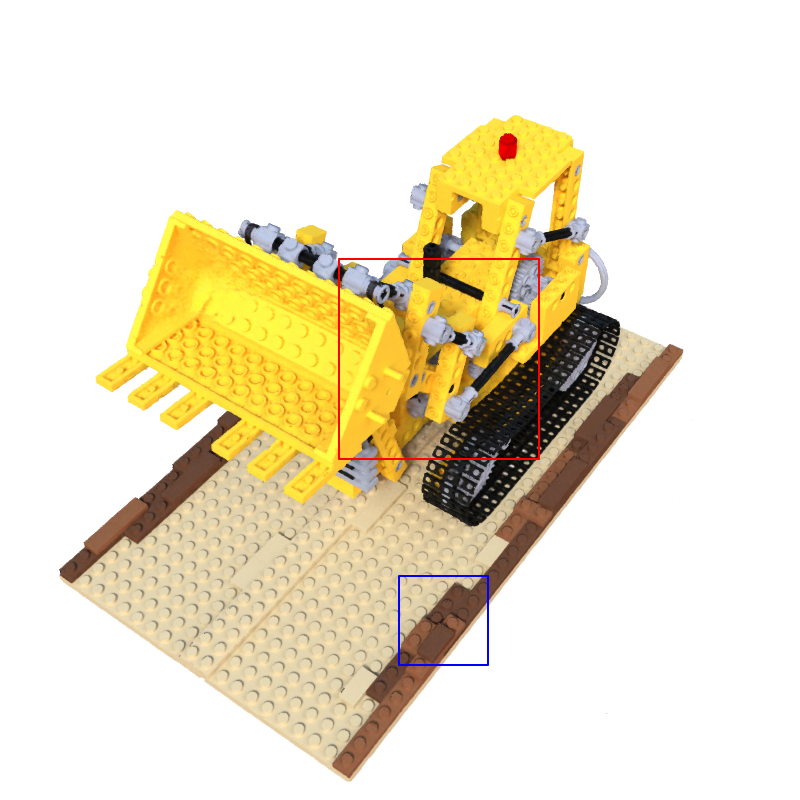}
         } \\
         \includegraphics[width=0.24\linewidth]{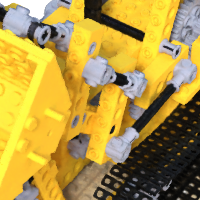} &
         \includegraphics[width=0.24\linewidth]{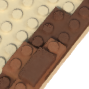} &
         \includegraphics[width=0.24\linewidth]{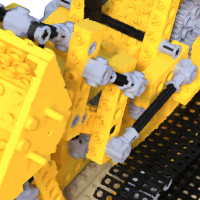} &
         \includegraphics[width=0.24\linewidth]{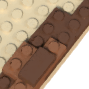} \\  
         \multicolumn{2}{c}{FP32 (32.61dB)} & \multicolumn{2}{c}{4-bit (32.61dB)} \\
    \end{tabular}
    \caption{Experiment using our 4-bit quantization model on SHACIRA~\cite{girish2023shacira} 3D NERF. Our model is  visually indistinguishable from the full precision model.}
    \label{shacira-result}
 \end{figure}

Our approach would allow to effectively compress any MLP-based INR. We tried our compression approach on SHACIRA~\cite{girish2023shacira}, which improves InstantNGP~\cite{muller2022instant}. We aim to prove that our quantization approach can be extended to other modalities. We applied 4-bits QAT to the MLP model in SHACIRA, keeping the latent space optimization as the original. 
We obtained $8\times$ model size reduction with zero degradation (\textit{32.61dB}). However, the actual model size reduction is small (from 1.96 MB to 1.82 MB) as the latent space size accounts for 90\% of the total size (around 1.81MB). Figure ~\ref{shacira-result} compares the qualitative results of the full precision and the quantized SHACIRA model.

We also provide samples in the supplementary material where we show 4-bit NeRFs~\cite{mildenhall2021nerf} without apparent loss.

%%%%%%%%%%%%%%%%%%%%%%%%%%%%%%%%%%%%%%%%%%%%%%%%%%
%%%% OFA
%%%%%%%%%%%%%%%%%%%%%%%%%%%%%%%%%%%%%%%%%%%%%%%%%%

%%%%%%%%%%%%%%%%%%%%%%%%%%%%%%%%%%%%%%%%%%%%%%%%%%

\section{Technical and Implementation Details}
We implement all the methods in PyTorch, using the author's implementations when available. We train all the models using the same environment with the Adam~\cite{kingma2014adam} optimizer, and we adapt the learning rate for each method's requirements. For instance MFN~\cite{fathony2020multiplicative} uses $0.01$ while SIREN~\cite{sitzmann2020implicit} uses $0.001$.

We use NVIDIA RTX 2080Ti and A100 (40GB and 80GB) cards. The models are optimized using the $\mathcal{L}_2$ reconstruction loss~\cite{tancik2020fourier, sitzmann2020implicit} to minimize the RGB image reconstruction error $\sum_{x,y} \| \mathcal{I}[x,y] - \phi (x,y) \|_{2}^{2} , ~~\forall (x,y) \in [H,W]$.  Note that due to the memory requirements of FHD images, the optimization is only possible on GPU cards with $>40$Gbs of VRAM.

\paragraph{Ablation Studies} In the supplementary, we provide ablation studies on the impact of layer quantization \ie which layers suffer the most.

\paragraph{Applications} Our method not only represent a theoretical contribution, ANIs allow to rethink content storage and transmission. Since we do not need to stream whole neural network to decode partial information, yet just a few layers, ANIs could have beneficial impact in remote sensing \ie satellite imagery transmission.

\paragraph{Limitations.} A clear limitation of using INRs for neural image compression is their stochastic nature and unstable training. Moreover, there is no practical way of predicting \emph{a priori} which INR model fits best the signal. On the other hand, Having the once-for-all alleviates this process, as a diverse array of PSNR/bpp ratios is readily available for serving.

%%%%%%%%%%%%%%%%%%%%%%%%%%%%%%%%%%%%%%%%%%%%%%%%%%

\newpage
\section{Conclusion}

In this work, we present a novel analysis on compressing neural representations. We also introduce Adaptive Neural Images (ANI), an efficient neural representation that enables adaptation to different inference or transmission requirements. We derive our ANI super-network using advanced once-for-all architecture search. To the best of our knowledge, we are the first approach to achieve successful 4-bit quantization of neural representations, establishing a new state-of-the-art. Moreover, this work provides the most complete benchmark for this task. Our work offers a transversal framework for developing compressed neural fields.

We refer the reader to our work ``Streaming Neural Images''\cite{conde2024streaming} for further studies on streaming neural images.

%%%%%%%%%%%%%%%%%%%%%%%%%%%%%%%%%%%%%%%%%%%%%%%%%%
% ---- Bibliography ----
%

\clearpage

\clearpage
\setcounter{page}{1}
\maketitlesupplementary
\vspace{10cm}

\setcounter{section}{0}
\setcounter{figure}{0}
\setcounter{table}{0}

\renewcommand{\thetable}{\Alph{table}}
\renewcommand{\thefigure}{\Alph{figure}}

%%%%%%%%%%%%%%%%%%%%%%%%%%%%%%%%%%%%%%%%%%%%%%%%%%%%%%%%%%%%%%%%%%%%%%%%

\section{Selecting the Neural Model}
\label{sec:inr}

We analyze the current \sota of INRs and their limitations. Although INRs are popular for image compression, there are not many benchmarks on well-known datasets.
We compare FourierNets~\cite{tancik2020fourier} (MLP with Positional Encoding), SIREN~\cite{sitzmann2020implicit}, MFN~\cite{fathony2020multiplicative}, Wire~\cite{saragadam2023wire} and DINER~\cite{xie2023diner} using well-known dataset for image processing. %We show a visual comparison in Figure~\ref{fig:baboon}.
We conduct an exhaustive study using the well-known Kodak dataset~\cite{kodak}. 

We present the results in Table~\ref{tab:kodak_inr}. From this experiment we can conclude: (i) the performance of an INR highly depends on the target image. (ii) Given an image, the performance of an INR can vary notably due to the random initialization -- for this reason we repeat each training 5 times with different random seeds. (iii) As we show in Figure~\ref{fig:train_evol}, the training --overfitting-- can be highly unstable depending on the model (\eg DINER~\cite{xie2023diner}). We provide more training samples in the appendix. The particular case of DINER~\cite{xie2023diner}~\emph{(CVPR '23)} suggests that using non-differentiable hash-tables difficulties training. 

Considering the experiment and conclusions, we decide to use mainly SIREN~\cite{sitzmann2020implicit} and MFN~\cite{fathony2020multiplicative} in our compression analysis because of the following reasons: (i) the models offer stable performance and good fidelity, (ii) do not require non-differentiable hash tables, (iii) their non-linear activations facilitate advanced quantization due to their mathematical properties.

\section{Additional Comparisons}

\begin{figure*}[t]
    \centering
    \begin{tabular}{c c}
         \includegraphics[width=0.45\linewidth]{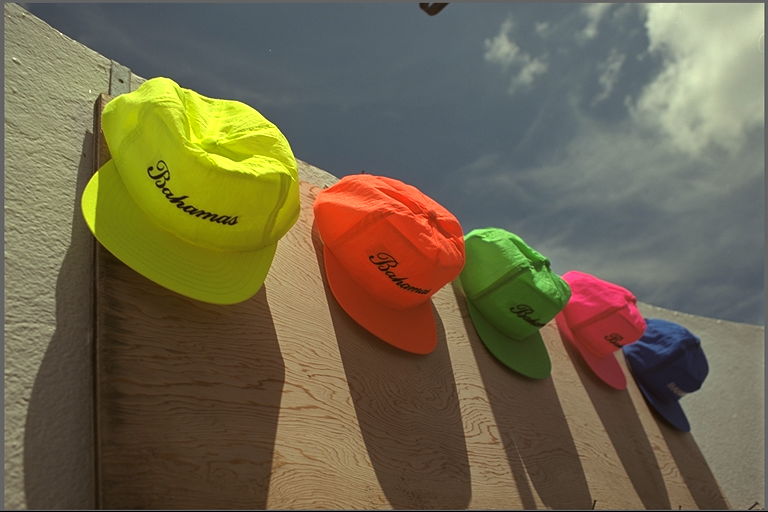} &
         \includegraphics[width=0.45\linewidth]{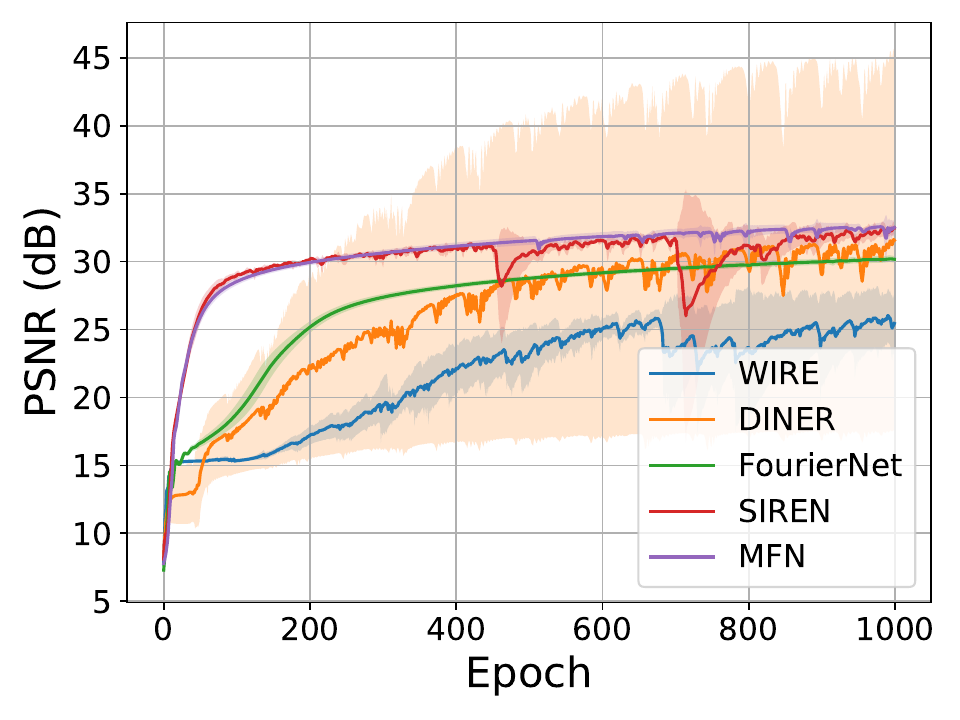} \\

         \includegraphics[width=0.45\linewidth]{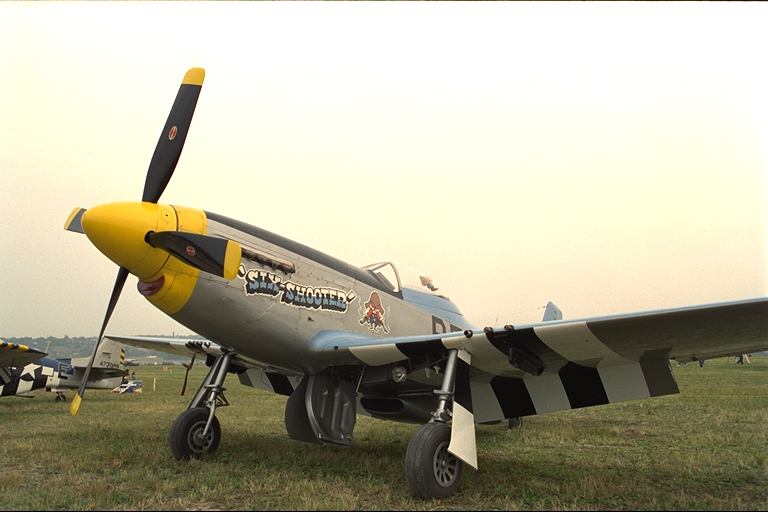} &
         \includegraphics[width=0.45\linewidth]{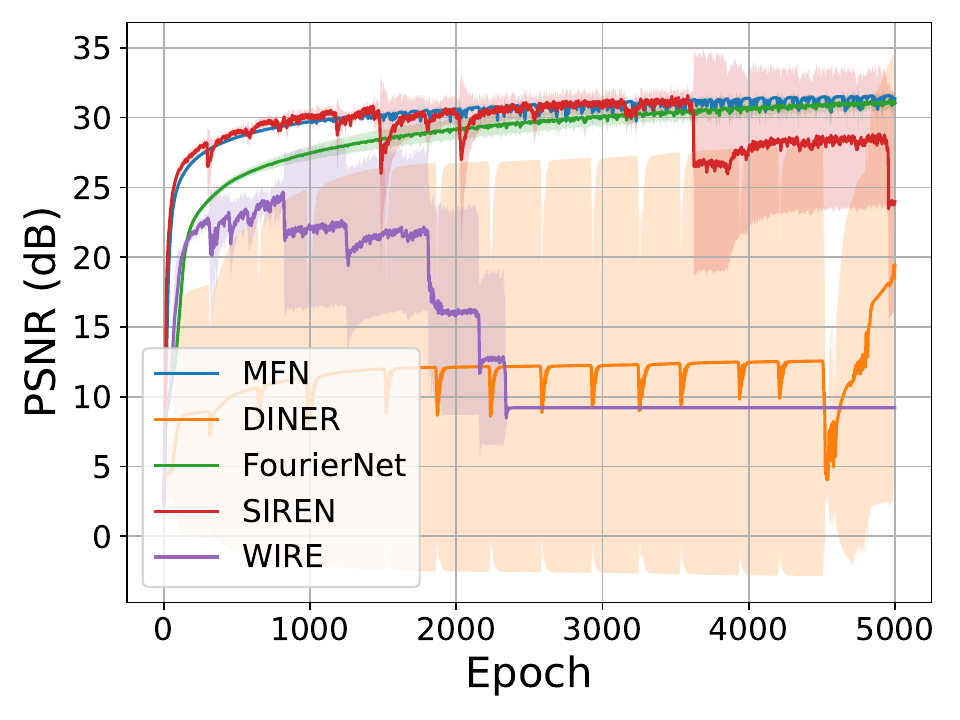} \\

         \includegraphics[width=0.45\linewidth]{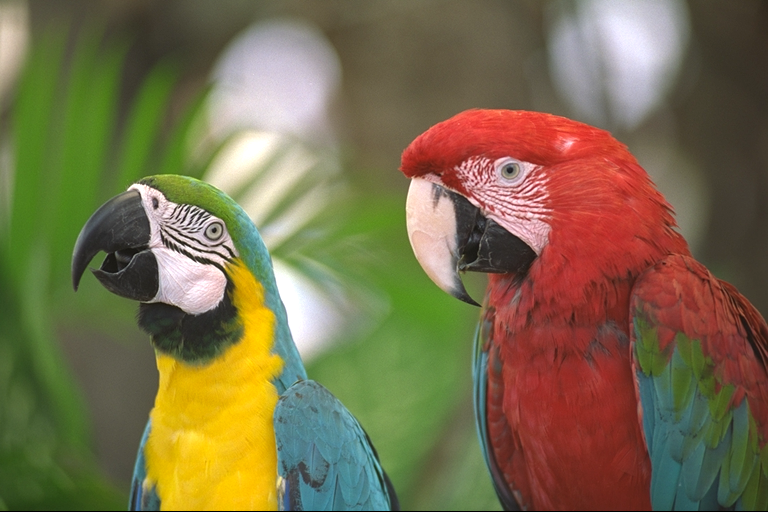} &
         \includegraphics[width=0.45\linewidth]{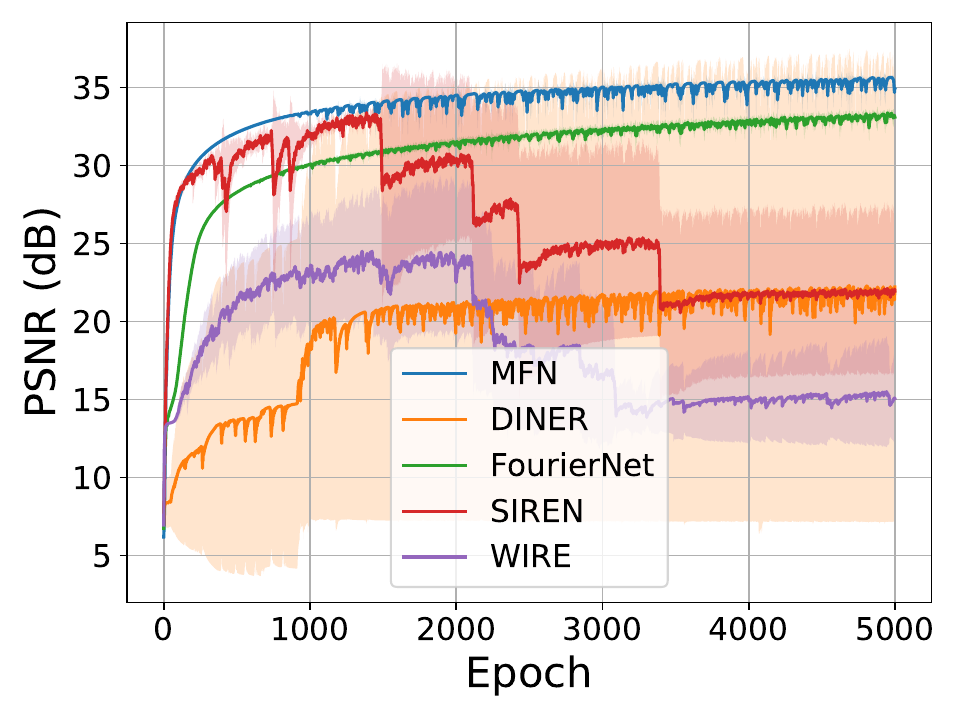} \\
    \end{tabular}
    \caption{Comparison of \sota INR backbones for neural image compression~\cite{sitzmann2020implicit, dupont2021coin, fathony2020multiplicative, xie2023diner, tancik2020fourier}. We can appreciate a great \emph{performance variability} depending on the target image. We show the average of 5 experiments (line) and their confidence intervals.}
    \label{fig:train-samples}
\end{figure*}

%\subsection{Training Variability}
%\section{Additional Training Samples}
We provide \textbf{additional training samples} using the Kodak dataset~\cite{kodak} in Figure~\ref{fig:train-samples}. We selected as backbones the most consistent models: SIREN~\cite{sitzmann2020implicit} and MFN~\cite{fathony2020multiplicative}. For each image we train the models five times and show the average (and std.). This is consistent across the dataset, and justifies our focus on SIREN~\cite{sitzmann2020implicit} and MFN~\cite{fathony2020multiplicative}.

\begin{figure*}[t]
    \centering
    \begin{tabular}{c c}
         \includegraphics[width=0.45\linewidth]{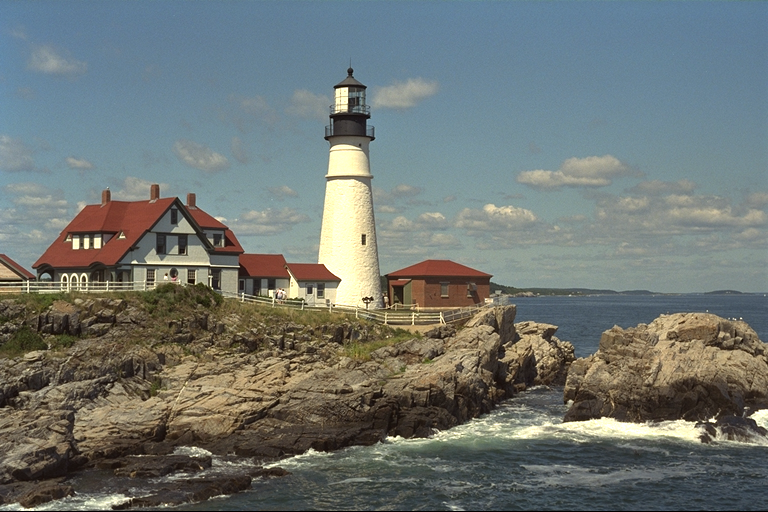} &
         \includegraphics[width=0.45\linewidth]{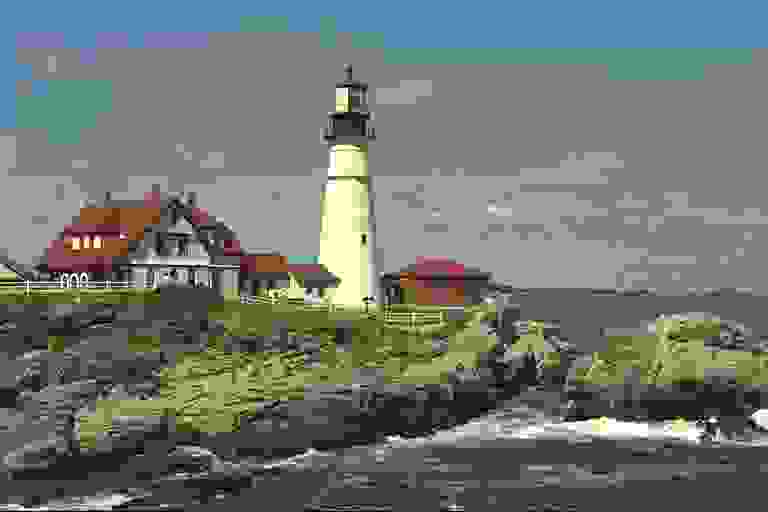} \\
         Original & JPEG~\cite{pennebaker1992jpeg} ($0.172$~bpp, $20.67$dB) \\
         \begin{tikzpicture}
            \node[anchor=south west,inner sep=0] (image) at (0,0) {\includegraphics[width=0.45\linewidth]{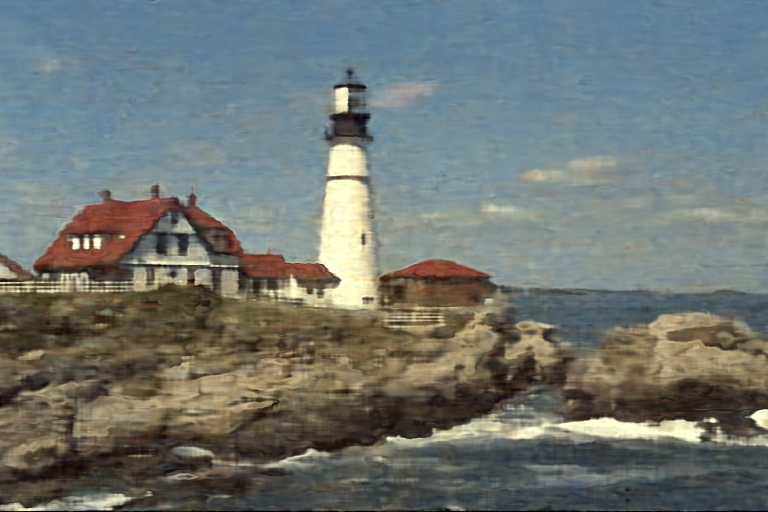}};
            \begin{scope}[x={(image.south east)},y={(image.north west)}]
                % Draw a red box (Adjust coordinates as needed)
                \draw[red,ultra thick] (0.2,0.2) rectangle (0.4,0.4);
                % The coordinates (0.3,0.3) and (0.6,0.6) define the lower left and upper right corners of the rectangle
                % These coordinates are relative to the image size (0,0) is bottom left, (1,1) is top right
            \end{scope}
        \end{tikzpicture}
        &
        \begin{tikzpicture}
            \node[anchor=south west,inner sep=0] (image) at (0,0) {\includegraphics[width=0.45\linewidth]{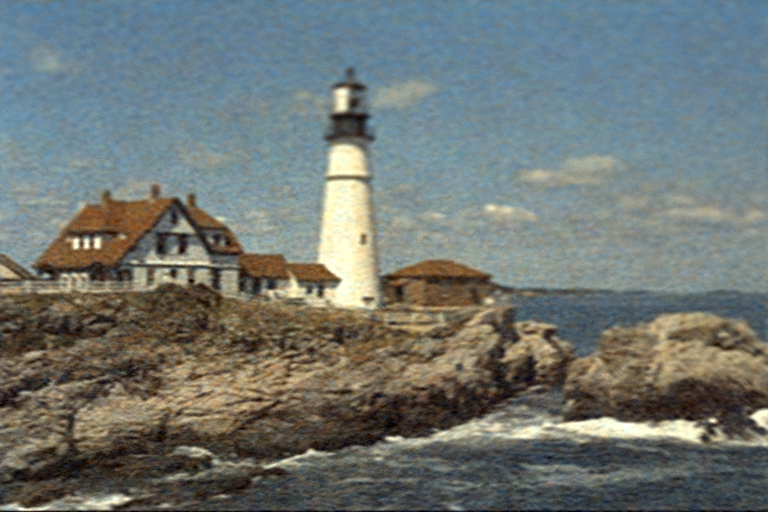}};
            \begin{scope}[x={(image.south east)},y={(image.north west)}]
                % Draw a red box (Adjust coordinates as needed)
                \draw[red,ultra thick] (0.2,0.2) rectangle (0.4,0.4);
                % The coordinates (0.3,0.3) and (0.6,0.6) define the lower left and upper right corners of the rectangle
                % These coordinates are relative to the image size (0,0) is bottom left, (1,1) is top right
            \end{scope}
        \end{tikzpicture}
          \\
         SIREN (COIN)~\cite{sitzmann2020implicit, strumpler2022implicit, dupont2021coin} ($24.43$dB) & ANI-MFN 8bits \emph{(ours)} ($25.69$dB) \\
    \end{tabular}
    \caption{\textbf{Comparison with previous works}~\cite{dupont2021coin, strumpler2022implicit} for neural image compression using SIREN~\cite{sitzmann2020implicit} as backbone. For a similar range of bpps $\approx0.17$, both methods provide similar compression capabilities. However, our novel approach ANI allows to adapt to different bpps, while previous works are fixed to a particular bpp. In the highlighted region we can appreciate how our method preserved more high-frequencies and details.}
    \label{fig:eccv_comp}
\end{figure*}

\begin{figure*}[t]
    \centering
    \begin{tabular}{c c c}
         \includegraphics[width=0.32\linewidth]{figures/y_eccv22/kodim21_01.jpg} &
         \includegraphics[width=0.32\linewidth]{figures/y_eccv22/OFAMFN_subnet_d2_w128.png} &
         \includegraphics[width=0.32\linewidth]{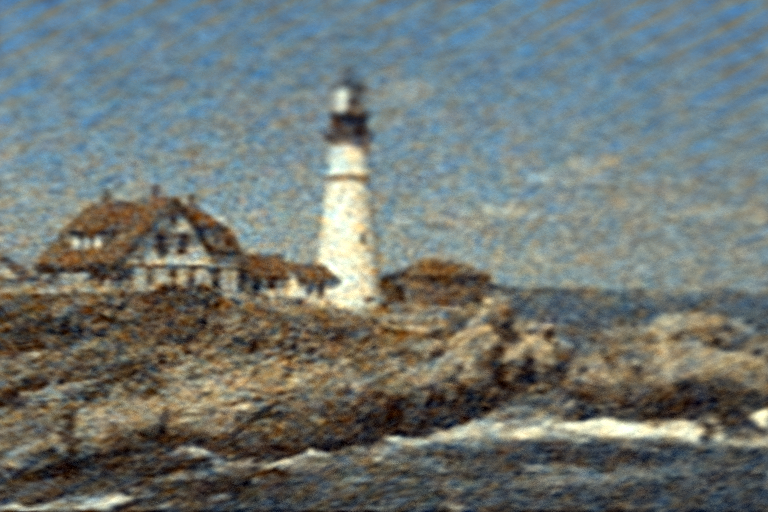} \\
         JPEG~\cite{pennebaker1992jpeg} & ANI-MFN $2\times128$ & ANI-MFN $2\times64$ \\
         $20.67$dB  & $25.69$dB & $23.36$dB   \\
    \end{tabular}
    \caption{\textbf{Comparison of artifacts under extreme compression ($<0.2$ bpp)}. Our ANI-MFN quantized models provide ``less harmful'' artifacts in comparison with JPEG, for instance we preserve color and structure. Our single model can be adapted to different bpp requirements. In comparison to JPEG ($0.17$bpp) our smallest model (2 layers, 64 neurons) can provide $0.06$bpps.}
    \label{fig:eccv_comp_artifacts}
\end{figure*}

In Figure~\ref{fig:eccv_comp} and Figure~\ref{fig:eccv_comp_artifacts} we provide high-resolution visual comparisons with previous works. Our approach can capture more high-frequencies and details under extreme compression setups ($<0.2$bpp). Moreover, \emph{our ANIs (adaptive neural images) can be adapted to different bpps}, unlike previous methods that are fixed to a particular bpp configuration (changing this would require re-training).

%%%%%%%%%%%%%%%%%%%%%%%%%%%%%%%%%%%%%%%%%%%%%%%%%%%%%%%%%%

\section{Sensitive layers analysis} 
\begin{figure*}[t]
    \centering
    \begin{tabular}{c c c c}
         \includegraphics[width=0.2\linewidth]{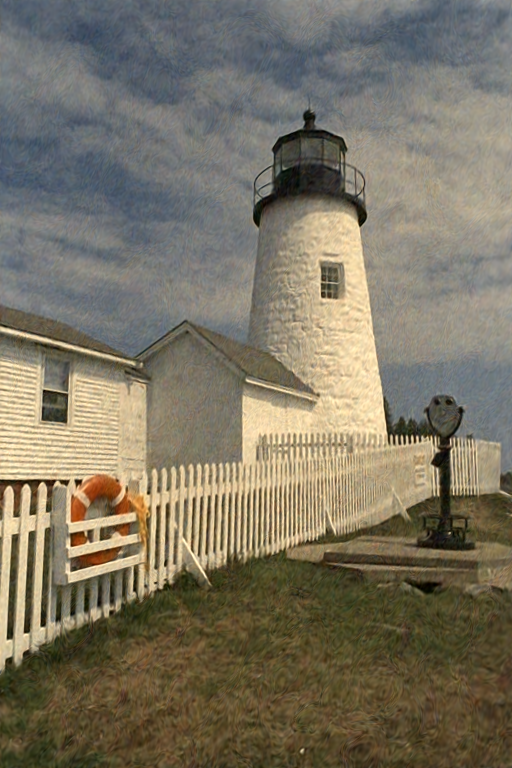} &
         \includegraphics[width=0.2\linewidth]{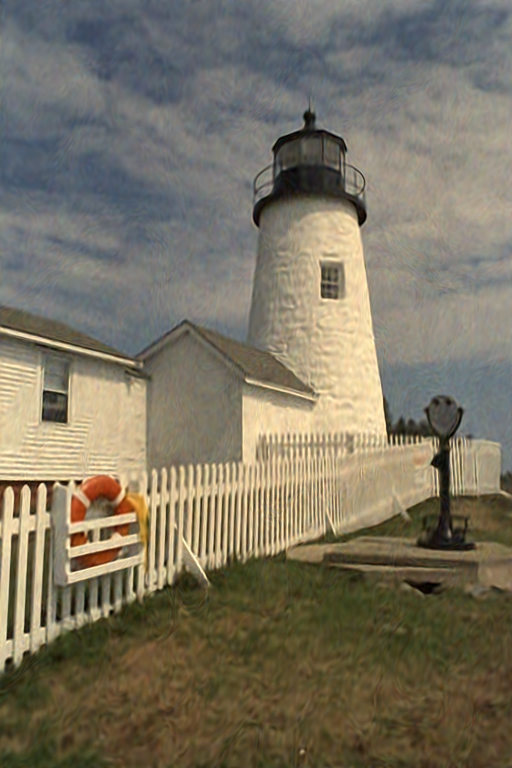} &
         \includegraphics[width=0.2\linewidth]{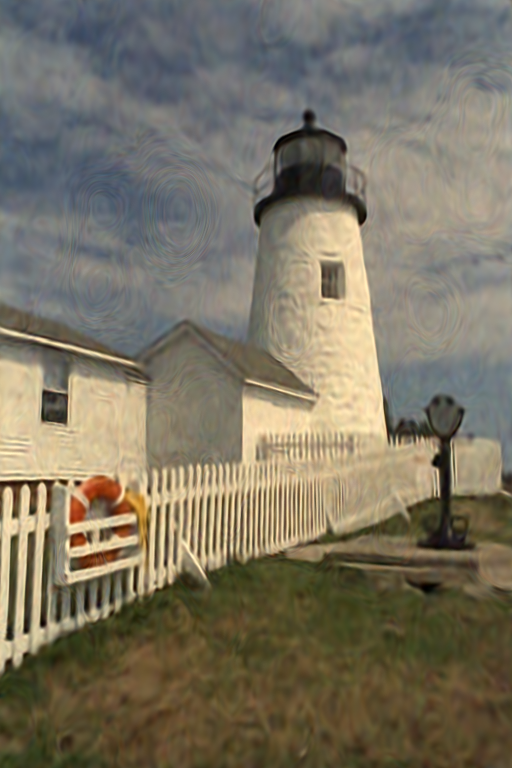} & 
         \includegraphics[width=0.2\linewidth]{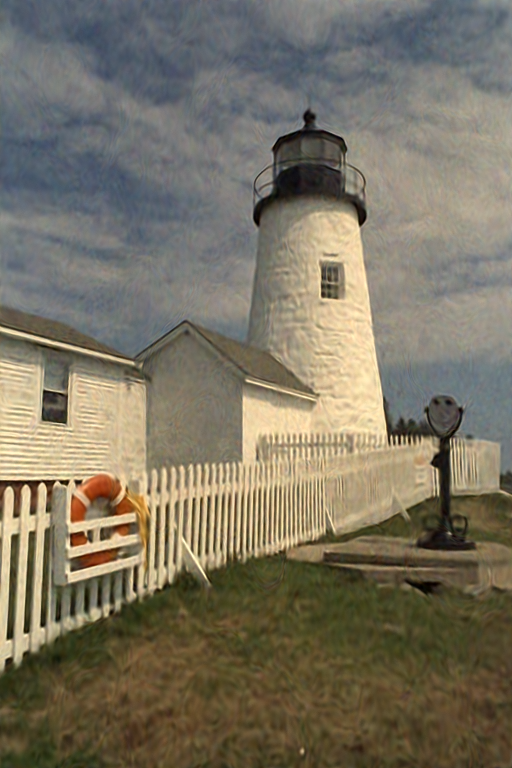} \\         SIREN~\cite{sitzmann2020implicit} & QAT 8-bit (default) & Quantize 1st layer & Quantize last layer \\
         \includegraphics[width=0.2\linewidth]{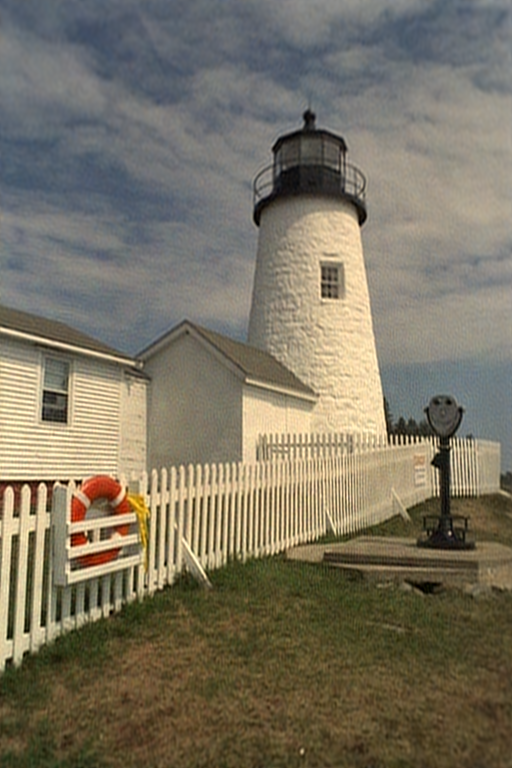} &
         \includegraphics[width=0.2\linewidth]{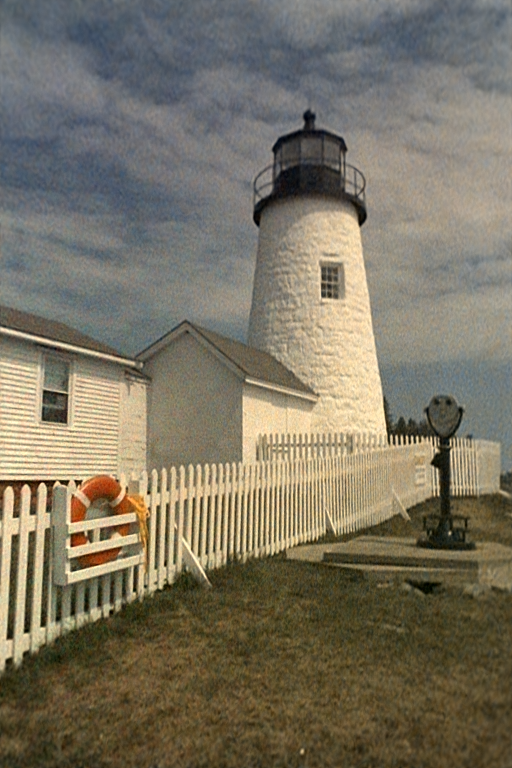} &
         \includegraphics[width=0.2\linewidth]{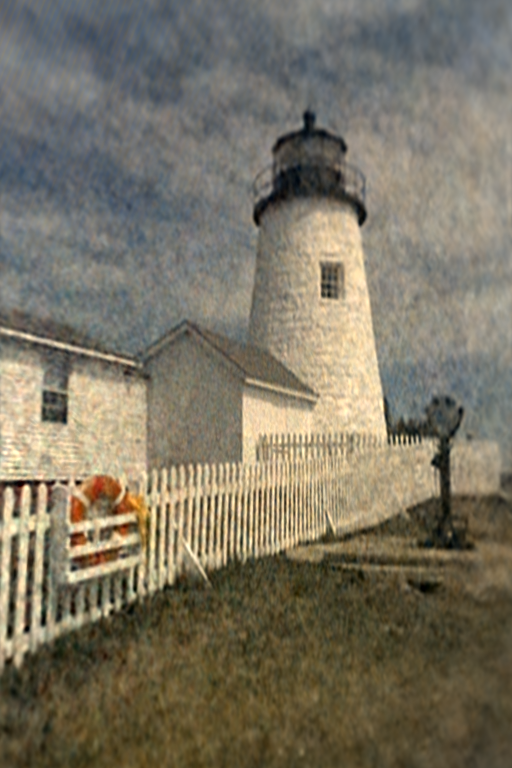} & 
         \includegraphics[width=0.2\linewidth]{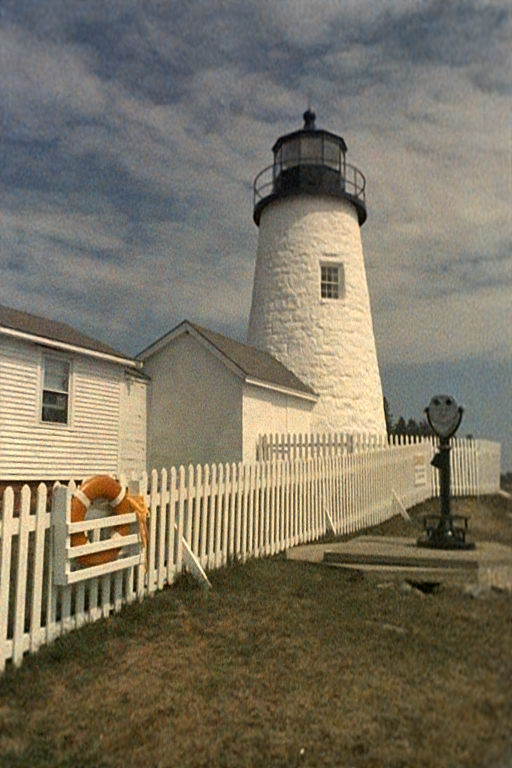} \\
         MFN~\cite{fathony2020multiplicative} & QAT 8-bit (default) & Quantize 1st layer & Quantize last layer \\
        
    \end{tabular}
    \caption{Ablation study for depth-wise quantization. We can appreciate the negative effects of quantizing the early layers, in contrast, quantizing the last layer does not affect much the fidelity. This denotes the importance of the first layers (or filters).}
    \label{fig:quant_first_last_layer_ablation}
\end{figure*}

Table~\ref{tab:per_layer_quantization} shows the impact of quantizing the first layer (filters for MFN) or the last layers. These layers are generally deemed sensitive. Quantizing the first layers drops the performance considerably with small compression in exchange, making it cost-effective. On the other hand, quantizing the final layer has barely any impact in the quality. See visual results in Figure~\ref{fig:quant_first_last_layer_ablation}.

\vspace{-2mm}
\paragraph{Theoretical INR Implications} Our empirical analysis provides evidence of the importance of early layers (or kernels) in the neural representation. The early layers contribute with the general basis and structural information, while the last layers provide signal-specific details.

%%%%%%%%%%%%%%%%%%%%%%%%%%%%%%%%%%%%%%%%%%%%%%%%%%%%%%%%%

%%%% INR X5 COMPARISON
\begin{comment}
\begin{figure*}[t]
    \centering
    \includegraphics[trim={0 8.5cm 0 7.5cm},clip, width=\linewidth]{figures/inr_comp/baboon_all.pdf}
    
    \vspace{-2mm}
    \caption{Results on the image \texttt{baboon} using state-of-the-art INRs~\cite{sitzmann2020implicit, tancik2020fourier, fathony2020multiplicative, saragadam2023wire, xie2023diner}.}
    \label{fig:baboon}
\end{figure*}
\end{comment}

\begin{table}[!ht]
    \centering
    \begin{minipage}[b]{\linewidth} 
        \centering
        \resizebox{0.9\columnwidth}{!}{
        \begin{tabular}{l c c c}
             \toprule
             \rowcolor{lgray} Method & Param.~(K) & PSNR~$\uparrow$ & SSIM~$\uparrow$ \\
             \midrule
             
             FourierNet~\cite{tancik2020fourier}  &  66.30 & 27.05$\pm$1.33 & 0.750 \\
             
             SIREN~\cite{sitzmann2020implicit}    &  66.82 & 27.98$\pm$2.73 & 0.782 \\
             
             MFN~\cite{fathony2020multiplicative} & 70.27  & 29.16$\pm$2.80 &  0.822 \\
             
             Wire~\cite{saragadam2023wire}        &  66.82 & 25.96$\pm$3.83 & 0.712 \\
             
             DINER~\cite{xie2023diner}            &  545.55 & 27.30$\pm$\textcolor{red}{15.2} & 0.750 \\
             \bottomrule
        \end{tabular}
        }
        \vspace{-2mm}
        \caption{Comparison of \sota INRs on the Kodak dataset~\cite{kodak}. We report the average PSNR ($\pm$ std.) and SSIM~\cite{wang2004ssim} over 5 runs. We observe high variability in DINER~\cite{xie2023diner}. The neural networks have $4$ layers, each with $128$ neurons. 
        }
        \label{tab:kodak_inr}
    \end{minipage}
    
    \begin{minipage}[b]{\linewidth}
        \centering
        \includegraphics[width=0.95\linewidth]{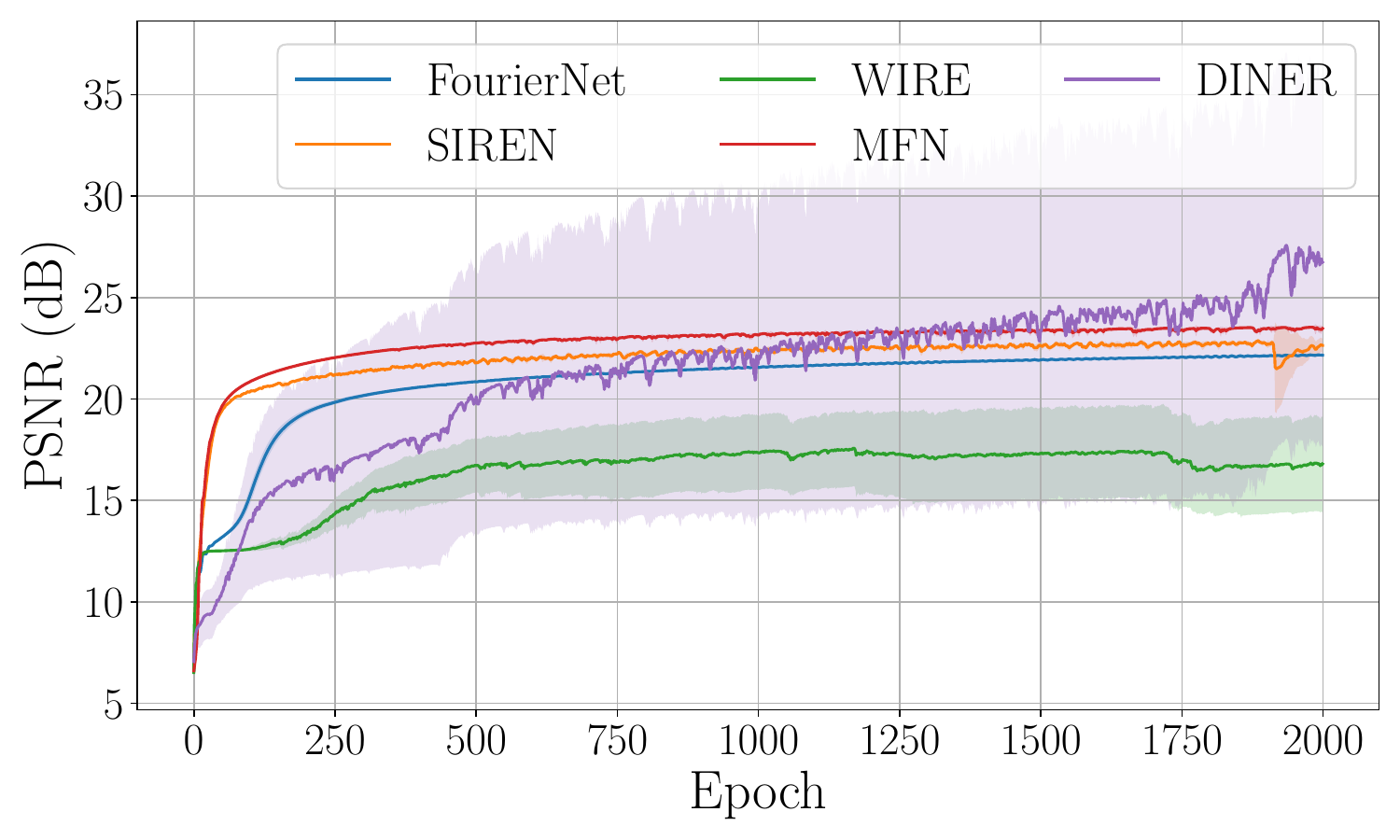} 
        \vspace{-4mm}
        \caption{Samples of training evolution of the different INR methods~\cite{tancik2020fourier, sitzmann2020implicit, fathony2020multiplicative, saragadam2023wire, xie2023diner}. This corresponds to \texttt{baboon}. We show the average results over 5 runs, and its confidence interval.} 
    \label{fig:train_evol}
    \end{minipage}
\end{table}

%%%%%%%%%%%%%%%%%%%%%%%%%%%%%%%%%%%%%%%%%%%%%%%%%%%%%%%%%

\begin{table}[!ht]
    \centering
    \resizebox{0.825\columnwidth}{!}{
        \begin{tabular}{l c c c}
             \toprule
             \rowcolor{lgray} Method & PSNR~$\uparrow$ & SSIM~$\uparrow$ & BPP~$\downarrow$\\
             \midrule
             SIREN~\cite{sitzmann2020implicit} & 27.73$\pm$0.14 & 0.771 & 1.833\\
             8-bit QAT default & 26.59$\pm$0.50 & 0.729 &  0.479\\
             + 8-bit first layer & 23.74$\pm$0.29 & 0.587 & 0.474 \\
             + 8-bit last layer & 26.25$\pm$0.54 & 0.715 & 0.471 \\
             \midrule
             MFN~\cite{fathony2020multiplicative} &  28.49$\pm$0.23 & 0.804  & 1.928 \\  
             8-bit QAT default & 28.72$\pm$0.05 & 0.782 &  0.572\\
             + 8-bit filters & 25.10$\pm$0.18 & 0.666 & 0.546  \\
             + 8-bit last layer & 28.42$\pm$0.12 & 0.764 & 0.565  \\ 
             \bottomrule
        \end{tabular}   
    }
    \vspace{-2mm}
    \caption{Impact of sensitive layers quantization. We use 4 layers and 128 neurons, and train for 2000 iterations.}
    \label{tab:per_layer_quantization}
\end{table}

%%%%%%%%%%%%%%%%%%%%%%%%%%%%%%%%%%%%%%%%%%%%%%%%%%%%%%%%%%
%% ADD TRAINING SAMPLES

%%%%%%%%%%%%%%%%%%%%%%%%%%%%%%%%%%%%%%%%%%%%%%%%%%%%%%%%%%
%%% ADD RESULTS

%%%%%%%%%%%%%%%%%%%%%%%%%%%%%%%%%%%%%%%%%%%%%%%%%%%%%%%%%%

%%%%%%%%%%%%%%%%%%%%%%%%%%%%%%%%%%%%%%%%%%%%%%%%%%%%%%%%%%

\clearpage

{
\small
\bibliographystyle{ieeenat_fullname}
\bibliography{main}
}

\end{document}